%% file: root.tex
\documentclass[lettersize,journal]{IEEEtran}
\usepackage[dvipsnames,table]{xcolor}
\usepackage{amsmath,amsfonts,amsthm}
\usepackage{algorithmic}
\usepackage{algorithm}
\usepackage{array}
\usepackage[caption=false,font=normalsize,labelfont=sf,textfont=sf]{subfig}
\usepackage{textcomp}
\usepackage{stfloats}
\usepackage{url}
\usepackage{verbatim}
\usepackage{graphicx}
\usepackage{cite}
\usepackage{hyperref}
\usepackage{multirow}
\usepackage[table]{xcolor}
\usepackage{siunitx}
\usepackage{mathtools}
\usepackage{bm}
\usepackage{bbm}
\usepackage{booktabs}
\usepackage{placeins}
\DeclareMathOperator*{\argmax}{arg\,max}

\usepackage{caption}
\makeatletter
\apptocmd{\@maketitle}{\input{figures/overview}}{}{}
\makeatother

\newtheorem*{remark}{Remark}

\begin{document}

\title{One Filter to Deploy Them All: Robust Safety for Quadrupedal Navigation in Unknown Environments}

\author{Albert Lin\textsuperscript{1,2}, Shuang Peng\textsuperscript{1}, %
and Somil Bansal\textsuperscript{2}%
\\
\textsuperscript{1}University of Southern California \quad \textsuperscript{2}Stanford University\\
Project Website: \href{https://sia-lab-git.github.io/One_Filter_to_Deploy_Them_All}{\texttt{https://sia-lab-git.github.io/One\_Filter\_to\_Deploy\_Them\_All}}
}

\maketitle

\begin{abstract}
As learning-based methods for legged robots rapidly grow in popularity, it is important that we can provide safety assurances efficiently across different controllers and environments.
Existing works either rely on \textit{a priori} knowledge of the environment and safety constraints to ensure system safety or provide assurances for a specific locomotion policy.
To address these limitations, we propose an observation-conditioned reachability-based (OCR) safety-filter framework.
Our key idea is to use an OCR value network (OCR-VN) that predicts the optimal control-theoretic safety value function for new failure regions and dynamic uncertainty during deployment time.
Specifically, the OCR-VN facilitates rapid safety adaptation through two key components: a LiDAR-based input that allows the dynamic construction of safe regions in light of new obstacles and a disturbance estimation module that accounts for dynamics uncertainty in the wild.
The predicted safety value function is used to construct an adaptive safety filter that overrides the nominal quadruped controller when necessary to maintain safety.
Through simulation studies and hardware experiments on a Unitree Go1 quadruped, we demonstrate that the proposed framework can automatically safeguard a wide range of hierarchical quadruped controllers, adapts to novel environments, and is robust to unmodeled dynamics \textit{without a priori access to the controllers or environments} - hence, ``One Filter to Deploy Them All''.
The experiment videos can be found on the project website.
\end{abstract}

\begin{IEEEkeywords}
Hamilton-Jacobi reachability analysis, safety filtering, adaptive safety, robust verification, safe legged locomotion.
\end{IEEEkeywords}

\input{sections/intro_somil}

\input{sections/related_works}

\input{tables/nomenclature_table}

\input{sections/problem_setup}

\input{sections/background}

\input{sections/approach}

\input{sections/simulation_experiments}

\input{sections/hardware_experiments}

\input{sections/limitations}

\input{sections/conclusion}

\section*{Acknowledgments}
This work is supported in part by a NASA Space Technology Graduate Research Opportunity, the NSF CAREER Program under award 2240163, and the DARPA ANSR program.

\bibliographystyle{IEEEtran}
\bibliography{bibliography/bansal_papers, bibliography/formal_safety_references, bibliography/opt_ctrl_and_dp, bibliography/reachability, bibliography/references,
bibliography/one_filter_references}

\newpage

\appendices
\input{appendices/proof}
\input{appendices/results}

\end{document}

%% file: figures/overview.tex
\centering
\includegraphics[width=1.0\textwidth]{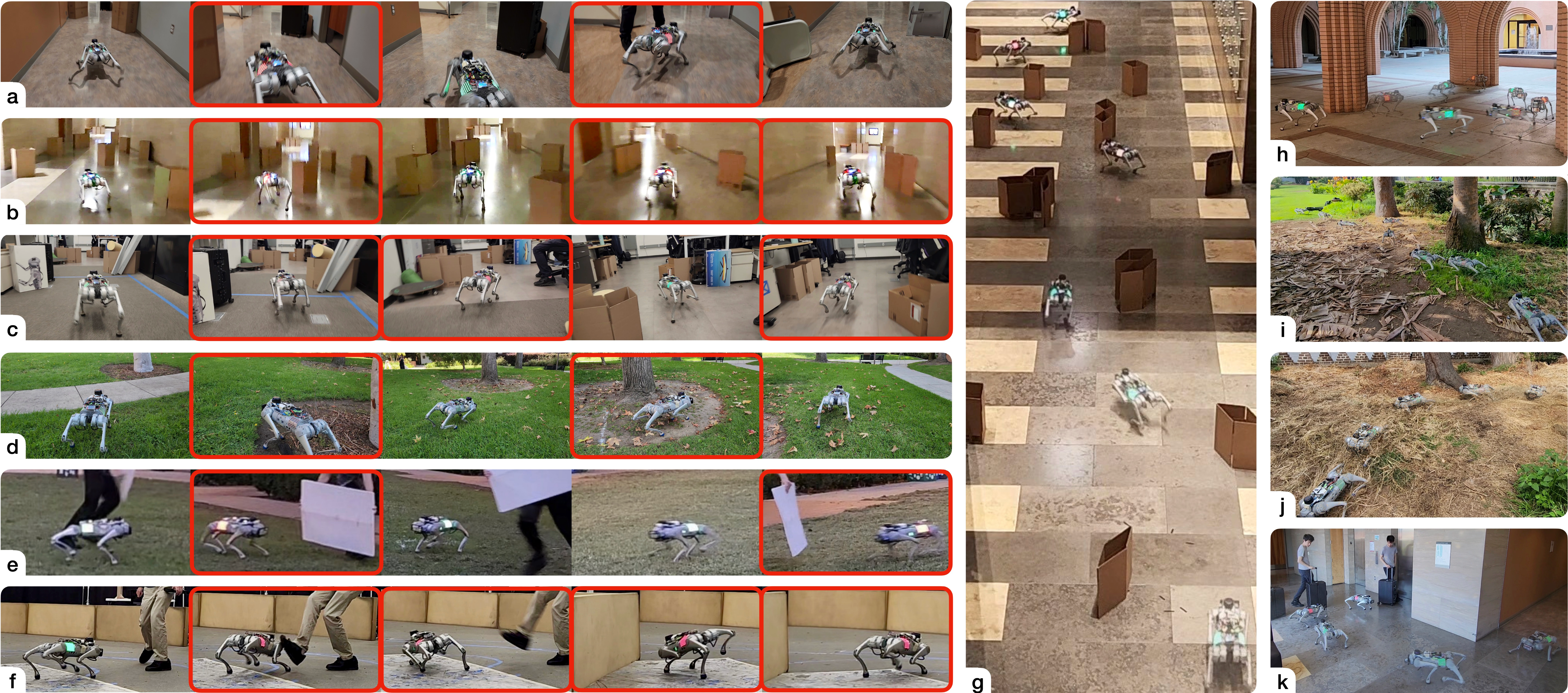}
\vspace{-0.5cm}
\captionof{figure}{
Our proposed observation-conditioned reachability-based (OCR) safety-filter framework automatically safeguards different controllers in diverse settings \textit{without a priori access to the controllers or environments}.
A trained OCR value network governs the switch between 
 nominal and {\color{BrickRed}filtered} control using an onboard LiDAR sensor.
The framework successfully safeguards a variety of high-level planners, including (a) learning-based, (c, f, k) model-based, (b, d, g, h, i, j) human teleoperated, and (e) naive planners, on top of different low-level locomotion policies, including (a, f, i, j, k) learning-based and (b, c, d, e, g, h) model-based policies.
Safety is maintained despite (a, b, c) narrow corridors, (d, i, j) rough terrains, (e, k) dynamic obstacles, (f) external disturbances, and (h) collision-seeking human teleoperation.
}
\vspace{-0.8cm}
\label{fig:overview}
\setcounter{figure}{0}

%% file: sections/intro_somil.tex
\section{Introduction}\label{sec:introduction}
Legged robots hold immense potential across diverse real-world applications, such as hazardous inspections ~\cite{HALDER2023105814, 10.1007/978-981-15-9460-1_18}, search and rescue missions ~\cite{li2023fabrication, cruz2023mixed}, entertainment ~\cite{gao2021design, lee2014human}, and public safety ~\cite{9316173}. 
A fundamental requirement in these scenarios is the ability to operate reliably in cluttered and \textit{a priori unknown} environments. 
However, achieving this reliability poses a significant challenge, as legged locomotion controllers must balance high performance with safety (e.g., avoiding collisions) during deployment. 
This work focuses on designing controllers that enable safe, collision-free quadrupedal locomotion while maintaining agility in novel environments.

Existing approaches to designing (safe) controllers for legged locomotion can be broadly categorized into model-based and reinforcement learning (RL)-based methods. 
Model-based methods provide provable safety guarantees using frameworks such as model predictive control (MPC), barrier functions, and reachability analysis \cite{ames2017first, grandia2021multi, 10610210, tayal2023safe, 10341987}. 
However, model mismatches and online computational burden limit the applicability of these approaches in the wild.
In contrast, RL-based controllers have demonstrated impressive agility in complex terrains and unstructured environments ~\cite{doi:10.1126/scirobotics.abc5986, doi:10.1126/scirobotics.adi7566, bellegarda2022visual, tan2018sim, 10.1145/3386569.3392433, shi2023terrain, 10610137, kumar2021rma, margolis2022walktheseways, 9561639, 10610137} 
However, these controllers typically prioritize agility, treating collision avoidance as a soft constraint during training, which can result in unsafe behaviors in cluttered or unseen environments.

To address these shortcomings, recent works have explored safety critics and backup policies to safeguard RL-based controllers~\cite{10611340, 10.1609/aaai.v33i01.33013387, pmlr-v164-zhao22a, 9290355, 9392290, HSU2023103811, he2024agile, bharadhwaj2021conservative}. 
These approaches precompute or learn safety critics that indicate when the nominal policy is deemed unsafe. 
While promising, existing methods often fail to ensure safety beyond the distribution of dynamics, environments, or locomotion policies encountered during training. 
Additionally, learning reliable backup policies for unknown environments remains a persistent challenge.

In this work, we propose an \underline{O}bservation-\underline{C}onditioned \underline{R}eachability (OCR) safety-filter framework, designed to integrate the agility of a nominal robot policy with robust safety in cluttered and unknown environments. 
Our key idea is to use an OCR Value Network (OCR-VN), which predicts the optimal control-theoretic safety value function for new failure regions and dynamic uncertainties encountered during deployment. 
Specifically, the proposed framework achieves this adaptability through two key components: first, it leverages an observation-based exteroceptive input (a LiDAR scan in this work) to dynamically adapt the safety value function directly from raw sensory inputs, enabling robust collision avoidance in different scenarios with onboard sensing and computation.
Second, it employs a disturbance estimation module to compute bounds on dynamics uncertainty (e.g., due to slippage, modeling inaccuracies, or low-level tracking errors) using recent state-action histories and adapts the robustness of the safety value function based on these bounds.

The OCR framework predicts when a nominal policy might violate safety and provides corrective control commands for the robot if necessary. 
A key strength of the proposed framework is its generality -- it can be deployed with a wide variety of nominal legged locomotion policies without requiring any retraining or policy-specific tuning. 
Additionally, we propose a Hamilton-Jacobi reachability-based method to train the OCR-VN, ensuring robust and efficient safety filtering.
We validate our approach through extensive simulations and real-world experiments on a Unitree Go1 quadruped, demonstrating that the OCR framework provides a reliable safety layer across multiple existing legged locomotion policies (both model-based and learning-based) and a variety of environments, without requiring prior knowledge of the specific policy or environment.
In summary, our key contributions are:
\begin{itemize}
    \item A reachability-based safety-filtering framework that ensures safety across diverse quadruped controllers and environments, without a priori access to the controller or the environment;
    \item An online adaptation mechanism that dynamically adapts the system safety to real-world environment variations and modeling uncertainties;
    \item Simulation and hardware experiments demonstrating the superior efficacy and robustness of the proposed approach in ensuring safe legged locomotion.
\end{itemize}

%% file: sections/related_works.tex
\section{Related Works}\label{sec:related_works}

\subsection{Safe Legged Locomotion}

\subsubsection{Model-Based Safety}
Traditional approaches for obstacle avoidance use collision-free motion planning techniques in the configuration space \cite{9196777, 9340701, https://doi.org/10.1002/rob.21974}.
They satisfy kinematic safety constraints but do not consider the dynamics of the system, limiting motions to slow quasi-static trajectories.
However, recent advances in agile locomotion and its applications have resulted in the need to consider dynamics.

Model-based approaches, such as model predictive control (MPC), use hand-designed or learned dynamics models to compute optimal maneuvers that are dynamically feasible and satisfy safety constraints \cite{9561326, 9812280, 9682571, 10341896, artplanner, teng2021safetyawareinformativemotionplanning, 9290355}.
Despite their impressive performance, such model-based approaches are generally computationally intensive for online settings and can run into safety feasibility issues, especially in cluttered obstacle environments.
Additionally, although they often perform well in settings that are captured accurately by their models, the safety guarantees are not robust to model mismatches that a robot might encounter in the wild \cite{8772165, doi:10.1126/scirobotics.abc5986}.

\subsubsection{RL-Based Safety}
Given the challenges associated with existing model-based approaches for agile locomotion, model-free RL-based approaches have emerged as popular alternatives.
RL-based approaches have found remarkable success in synthesizing efficient and robust locomotion in the real world, especially as the availability of high-fidelity simulators has increased \cite{makoviychuk2021isaac, rudin2021learning}.
They are well-suited to handle complex high-dimensional systems, multimodal feedback signals, and difficult-to-specify task objectives \cite{rlsurveyrobotics}.

Previous studies have optimized locomotion policies for specific skills such as agility \cite{doi:10.1177/02783649231224053, doi:10.1126/scirobotics.aau5872}, resilience \cite{9779429, doi:10.1126/scirobotics.abb2174, doi:10.1126/scirobotics.abk2822}, and difficult terrain traversal \cite{kumar2021rma, doi:10.1126/scirobotics.abc5986, zhuang2023robot, zhang2023learning, doi:10.1126/scirobotics.adh5401, cheng2023parkour, margolis2022walktheseways, Miki_2022, wang2023learning, agarwal2023legged}.
However, these works typically focus on maximizing agility without regard to safe navigation.
These methods can be combined with high-level collision-free planners; however,
they suffer from the aforementioned limitations of model-based controllers and restricted mobility \cite{Wellhausen_2023, 10611254, 10531249, he2024agile}.

Other works consider safe navigation during the learning process by including a large collision penalty in the reward function to incentivize collision-avoidance \cite{9385894, yang2022learning, 10611254, 9981198, zhang2024learningagilelocomotionrisky, doi:10.1126/scirobotics.adi7566, 10161302, kareer2023vinl, truong2022rethinking, 10328058}.
Unfortunately, there are no formal guarantees of safety, and the synthesized locomotion policies can degrade in safety when transferred to the real world due to a distribution shift away from the environments seen during training.

\subsubsection{Certificate-Based Safety}

In order to provide rigorous safety assurances, many works have proposed certificate-based safety methods within both model-based \cite{9561326, 9812280, 10341896, teng2021safetyawareinformativemotionplanning, 9290355} and RL-based \cite{9982038, 10611340, 10.1609/aaai.v33i01.33013387, pmlr-v164-zhao22a, 9392290, HSU2023103811, he2024agile, bharadhwaj2021conservative} frameworks, most often using control barrier functions or reachability-based value functions.
These methods are typically reliant on the offline availability of a certificate function or dynamics, which limits their applicability to complex real-world systems.

Some recent works learn adaptive safety certificates and recovery policies to ensure safety at runtime \cite{9982038, 10611340, 10.1609/aaai.v33i01.33013387, pmlr-v164-zhao22a, 9392290, HSU2023103811, he2024agile}.
He et al. \cite{he2024agile} propose Agile But Safe (ABS), an approach that co-designs performance and control-theoretic safety controllers via RL and switches between them as necessary to maintain safety.
Since the computed safety assurances are policy-dependent, they can be suboptimal and overly conservative, depending on the quality of the nominal policy.
Moreover, the safety controller needs to be retrained as the policy is fine-tuned, which can be cumbersome and time-consuming.
Additionally, since approaches like ABS lack active mechanisms to account for unmodeled dynamics during deployment (e.g., slippage), the computed safety assurances are valid only within the distribution of environments and locomotion policies seen during training.
These limitations are especially relevant due to the rising popularity of diverse learning-based policies, whose safety assurances must be updated as they are fine-tuned.

Our work most closely aligns with certificate-based safety methods like ABS, but we address several existing limitations.
Instead of computing the policy-conditioned safety function, we compute the optimal control-theoretic safety controller via Hamilton-Jacobi (HJ) reachability analysis.
This enables us to construct a safety filter for any nominal controller without needing to recompute the safety assurances.
We discard the need for high-fidelity simulators by using a reduced-order system and robustly handle the model gap as an adversarial disturbance in the dynamics.
By estimating and adapting to disturbance bounds during deployment, our proposed framework is able to ensure safety more robustly across a range of settings and policies compared to previous works.

\subsection{Reachability-Based Safety Filters}
This work extends the class of Hamilton-Jacobi (HJ) reachability-based filters \cite{10665911, 10266799, hsu2023safety} which ensure safety at deployment by overriding a nominal controller when necessary to preserve safety.
Due to the computational burden, traditional reachability-based safety filters are typically constructed offline for systems assumed to be known a priori \cite{bansal2017hamilton}.
In the case of quadrupedal navigation, a quadruped using a specific locomotion policy is often abstracted as a reduced-order system for computational tractability.
The dynamics of the system depend on the locomotion policy; thus, reachability-based filters will not readily safeguard different locomotion policies.
Additionally, it is challenging to adapt safety assurances to different obstacle configuration and terrain properties due to the computational burden \cite{bansal2019combining, 10160554}.
This work overcomes these limitations by distilling the safety solutions for a wide range of system settings into an Observation-Conditioned Reachability-based Value Network (OCR-VN).
During deployment, the robot adapts by querying the trained OCR-VN with the current state, control, and observation history.

%% file: tables/nomenclature_table.tex
\begin{table}[!h]
\caption{Nomenclature}\label{tab:nomenclature}
\centering
\begin{tabular}{p{0.26\linewidth}p{0.64\linewidth}}

\textbf{Symbol} & \textbf{Definition.} \\

\hline

$t$, $T$ & Time, time horizon. \\

$e \in \mathcal{E}$ & Environment. \\

$x \in \mathcal{X}$ & State. \\

$u \in \mathcal{U}$ & Control input. \\

$d \in \mathcal{D}$ & Disturbance input. \\

$f$ & System dynamics. \\

$\pi$, $\pi^\text{high}$, $\pi^\text{low}$ & Hierarchical, high-level, and low-level policies. \\

$\xi^{\pi}_{x,t}(\tau)$ & State achieved at time $\tau$ by starting at initial state $x$ at time $t$ and applying policy $\pi$ over $[t, \tau]$. \\

$\mathcal{F}$, $l$ & Failure set and function. \\

$o \in \mathcal{O}$ & Observation. \\

$x_r \in \mathcal{X}_r \subseteq \mathcal{X}$, $f_r$ & Reduced-order state and dynamics. \\

$\omega = \left( v, w \right)$ & Twist $\omega$ consisting of velocity $v$ and yaw rate $w$. \\

$\bar{d}_{p_x,p_y}$, $\bar{d}_{p_\theta}$ & Disturbance bound in $p_x, p_y$ and in $p_\theta$. \\

$\bar{d}_r = [\bar{d}_{p_{x},p_{y}}, \bar{d}_{p_{\theta}}]$ & Disturbance bound for reduced-order system. \\

$V$, $V_\psi$ & Ground-truth and learned value functions. \\

\\

\textbf{Abbreviation} & \textbf{Definition.} \\

\hline

OCR & Observation-conditioned reachability. \\

ABS & Agile But Safe \cite{he2024agile}. \\

WTW & Walk-These-Ways \cite{margolis2022walktheseways}. \\

MPC & Model predictive control policy by Unitree \cite{unitree_ros}. \\

PS & Predictive sampling-based planner. \\

NVE & Naive planner. \\

HMN & Human-teleoperated planner. \\

\end{tabular}
\vspace{-1.2em}
\end{table}

%% file: sections/problem_setup.tex
\section{Problem Setup}\label{sec:problem_setup}

See Table \ref{tab:nomenclature} for notation. In this work, we are interested in ensuring the safety of a quadruped robot in an \textit{a priori unknown} environment $e \in \mathcal{E}$.
Here, $e$ contains all the information needed to inform the effects of the environment on dynamics, as well as failure regions.
For example, $e$ can include the terrain geometry, friction coefficients, and obstacle locations.

We model the quadruped as a nonlinear dynamical system with state $x \in \mathcal{X}$, control $u \in \mathcal{U}$, and dynamics $\dot{x} = f(x, u; e)$ governing how $x$ evolves over time until a final time horizon $T$.
The dynamics are also affected by the environment $e$, e.g., by the effects of a slippery floor.
We denote the robot observations from proprioception and/or exteroception as $o^e_x=h(x; e) \in \mathcal{O}$.
For exteroception, we primarily deal with LiDAR scans in this work, though other sensors can also be used.
We denote the set of failure states as $\mathcal{F}^e \subseteq \mathcal{X}$ (e.g., collision states) which the robot is not allowed to enter.
The failure set can be represented by the zero-sublevel set of a Lipschitz-continuous function $l^e: \mathcal{X} \rightarrow \mathbb{R}$, i.e., $x \in \mathcal{F}^e \Leftrightarrow l^e(x) \leq 0$.
Note that $\mathcal{F}^e, l^e$ are also functions of $e$.

Let $\pi_\text{nom}$ denote a hierarchical nominal policy for the quadruped that takes the system history and outputs the robot control.
We assume that $\pi_\text{nom}$ consists of a high-level planner, $\pi_\text{nom}^\text{high}$, that provides twist commands $\omega \coloneqq (v, w)$ consisting of a forward velocity $v$ and a yaw rate $w$.
These twist commands are tracked by a low-level locomotion policy, $\pi_\text{nom}^\text{low}$, e.g., an RL-based policy \cite{doi:10.1177/02783649231224053, margolis2022walktheseways, he2024agile} or an MPC-based policy \cite{unitree_ros, 9340701}, that ultimately provides control inputs $u$ for the robot.
This architecture is popular in the legged robotics literature where $\pi_\text{nom}^\text{high}$ is typically designed for collision avoidance and navigation, and $\pi_\text{nom}^\text{low}$ could be an agile locomotion policy that can handle different terrains that the robot might encounter in the wild.

Let $\xi^{\pi}_{x, e, t}(\tau)$ denote the state achieved at time $\tau \in [t, T]$ by starting at initial state $x$ at time $t$ and applying the control policy $\pi$ over $[t, \tau]$ in environment $e$.
Our goal is to compute a safe policy $\pi_\text{safe}$ that ensures that the quadruped remains outside of the failure set at all times, i.e., $\pi_\text{safe}: \forall \tau \in [0, T], \xi^{\pi_\text{safe}}_{x, e, 0}(\tau) \notin \mathcal{F}^e$, while preserving the underlying performance of $\pi_\text{nom}$ to the extent possible.
The key challenge in designing such a policy is that the robot environment (and hence the failure set and the robot dynamics) is not known beforehand, necessitating a real-time update of the safety policy with the environment.
A second challenge stems from the fact that we want the safety framework to be agnostic to different nominal policies.

%% file: sections/background.tex
\section{Background}\label{sec:background}

\subsection{Hamilton-Jacobi Reachability}

Our proposed framework builds upon Hamilton-Jacobi (HJ) reachability analysis, which is a popular formal verification tool for computing safety guarantees for general nonlinear dynamical systems \cite{lygeros2004reachability, mitchell2005time}.
For reachability analysis, we will consider a more general form of dynamics $\dot{x} = f(x, u, d)$, where $d \in \mathcal{D}$ represents the disturbance.
Later on, in our work, we will use $d$ to model potential uncertainty in the system dynamics model.
We also omit the dependence on environment for now for brevity purposes.

HJ reachability analysis is concerned with computing the system's initial-time Backward Reachable Tube, which we denote as $\text{BRT}$.
We define $\text{BRT}$ as the set of all initial states $x \in \mathcal{X}$ starting from which, for all control signals $\mathbf{u}(\cdot)$, there exists a disturbance signal $\mathbf{d}(\cdot)$ such that the system will inevitably enter the failure set $\mathcal{F}$ within the time horizon $[0, T]$:
\begin{equation}
    \text{BRT} \coloneq \{ x \in \mathcal{X}: \forall \mathbf{u}(\cdot), \exists \mathbf{d}(\cdot),\exists \tau \in [0, T], \xi^{\mathbf{u}(\cdot),\mathbf{d}(\cdot)}_{x,0}(\tau) \in \mathcal{F} \}.
\end{equation}
By ensuring that the system remains outside of $\text{BRT}$, we guarantee system safety for the time horizon $T$.

In HJ reachability, computing $\text{BRT}$ is formulated as a robust optimal control problem.
First, we implicitly represent the failure set $\mathcal{F}$ by a failure function $l(x)$ whose zero-sublevel set yields $\mathcal{F}$: $\mathcal{F} = \{ x \in \mathcal{X}: l(x) \leq 0 \}$.
$l(x)$ is commonly the signed distance function to $\mathcal{F}$.
Next, we define the cost function corresponding to a control signal $\mathbf{u}(\cdot)$ and disturbance signal $\mathbf{d}(\cdot)$ to be the minimum of $l(x)$ over the trajectory starting from state $x$ and time $t$:
\begin{equation}
    J_{\mathbf{u}(\cdot), \mathbf{d}(\cdot)}(x, t) \coloneq \min_{\tau \in [t, T]}l(\xi^{\mathbf{u}(\cdot),\mathbf{d}(\cdot)}_{x,t}(\tau)).
\end{equation}
Since the control aims to avoid $\mathcal{F}$ under worst-case disturbance, the value function corresponding to this robust optimal control problem is:
\begin{equation}
    \label{eq:V}
    V(x, t) \coloneq \max_{\mathbf{u}(\cdot)} \min_{\mathbf{d}(\cdot)} J_{\mathbf{u}(\cdot), \mathbf{d}(\cdot)}(x, t).
\end{equation}
By defining our optimal control problem in this way, we can easily recover $\text{BRT}$ using the value function.
The value function being nonpositive implies that the failure function is nonpositive somewhere along the optimal trajectory, or in other words, that the system will inevitably enter $\mathcal{F}$.
Conversely, the value function being positive implies that there exists a control signal that will prevent the system from entering $\mathcal{F}$ even under the worst-case disturbance signal.
Thus, $\text{BRT}$ is computed as the zero-sublevel set of the initial-time value function:
\begin{equation}
    \label{eq:BRT}
    \text{BRT} = \{ x \in \mathcal{X}: V(x, 0) \leq 0 \}.
\end{equation}
The value function in Equation \eqref{eq:V} can be computed using dynamic programming, resulting in the following final value Hamilton-Jacobi-Isaacs Variational Inequality (HJI-VI):
\begin{align}
    \min\{ D_{t}V(x, t) + H(x, t, \nabla V(x, t)), l(x) - V(x, t) \} = 0, \nonumber \\
    V(x, T) = l(x), \quad \forall t \in [0, T].
\end{align}
$D_{t}V(x, t)$ and $\nabla V(x, t)$ represent the temporal derivative and spatial gradient of the value function $V(x, t)$, respectively.
The Hamiltonian $H(x, t, \nabla V(x, t))$ encodes how the control and disturbance interact with the system dynamics:
\begin{equation}
    H(x, t, \nabla V(x, t)) \coloneq \max_{u \in \mathcal{U}} \min_{d \in \mathcal{D}} \nabla V(x, t) \cdot f(x, u, d).
\end{equation}
The value function in Equation \eqref{eq:V} also induces the optimal safety controller:
\begin{equation}
    \label{eq:u*}
    \mathbf{u}^*(x, t) \coloneq \text{arg}\max_{u \in \mathcal{U}} \min_{d \in \mathcal{D}} \nabla V(x, t) \cdot f(x, u, d).
\end{equation}
Intuitively, the optimal safety controller aligns the system dynamics in the direction of the value function's gradients, thus steering the system towards higher-value states, i.e., away from $\mathcal{F}$.
An important result of HJ reachability theory is that safety is guaranteed despite worst-case disturbances if the system starts outside of $\text{BRT}$ and applies the control in Equation \eqref{eq:u*} at the $\text{BRT}$ boundary \cite{10665911}.

\subsection{HJ Reachability-Based Safety Filtering}\label{sec:slr}

In this work, we will use HJ reachability analysis to maintain the safety of the quadruped robot during deployment via HJ reachability-based safety filtering, where the computed safety value function is used to construct a safety filter that guarantees system safety.
While there are many different types of safety filters that can be constructed, we use the smooth least-restrictive safety filter, which aims to maximally preserve the underlying performance of a given nominal policy $\pi_\text{nom}$ by intervening as seldomly and as lightly as possible \cite{10665911}.
When outside of the system $\text{BRT}$, the smooth least-restrictive filter $\pi_\text{safe}$ outputs the nominal control.
At the boundary of the $\text{BRT}$, $\pi_\text{safe}$ outputs a safe control as close as possible to the nominal control by solving a quadratic program (QP).
\begin{equation} \label{eq:slr}
    \pi_\text{safe} (x, t) = 
    \begin{cases}
        \pi_\text{nom}(x, t), & V(x, t) > 0 \\
        \pi_\text{QP}(x, t), & V(x, t) = 0,
    \end{cases}
\end{equation}
where $\pi_\text{QP}(x, t)$ is obtained by solving:
\begin{equation}\label{eq:slr_qp}
    \begin{aligned}
        &\text{arg}\min_{u \in \mathcal{U}} || u - \pi_\text{nom}(x, t) ||_2^2 \\
        &\text{s.t. } D_{t}V(x, t) + \min_{d \in \mathcal{D}} \nabla V(x, t) \cdot f(x, u, d) = 0.
    \end{aligned}
\end{equation}
Intuitively, the constraint in \eqref{eq:slr_qp} enforces safe control of the system at the $\text{BRT}$ boundary, where the system safety could be in jeopardy.
It has been shown that the filter in \ref{eq:slr} is guaranteed to maintain system safety under worst-case disturbances as long as the system starts outside of the $\text{BRT}$ \cite{10665911}.
In the next section, we propose to construct an adaptive version of this safety filter by using an Observation-Conditioned Reachability-based Value Network (OCR-VN) that predicts the safety value function for new failure regions and dynamic uncertainties encountered during deployment.

%% file: sections/approach.tex
\section{Approach}\label{sec:approach}

\input{figures/framework}

The main difficulty in directly applying traditional HJ reachability methods in Section \ref{sec:background} to the problem in Section \ref{sec:problem_setup} is that the system dynamics $f(x, u; e)$ and the failure set $\mathcal{F}^e$ are functions of the unknown environment $e$.
Thus, a safety controller must be able to adapt to both \textit{dynamics uncertainty} and \textit{environment uncertainty}.
To overcome this issue, we propose the Observation-Conditioned Reachability (OCR) safety-filter framework.

The key idea behind our framework is to use an OCR Value Network (OCR-VN) that predicts the optimal safety value function for new dynamics $f(x, u; e)$ and failure regions $\mathcal{F}^e$ using the most recent system and observation history.
The OCR-VN facilitates rapid adaptivity through two key components: an observation-based input to the network (see Section \ref{sec:network}) and a disturbance estimation module (see Section \ref{sec:estimator}).
The observation-based input allows the OCR-VN to perceive new obstacles in the environment and dynamically adapt the value function to safeguard against them; whereas the disturbance estimation module allows the OCR-VN to tune the degree of uncertainty that is present within the dynamics model based on the most recent system history, and correspondingly adapt the safety value function to this dynamics uncertainty.
The resultant safety value function is then used to filter the nominal policy to ensure safety while maintaining the agility of the underlying policy.
Ultimately, our safety framework provides high-level twist commands, which, when combined with the low-level nominal policy, maintains the robot's safety.

The OCR framework can be divided into two distinct phases: training and deployment.
These are illustrated in Figure \ref{fig:framework}.
During the training phase, we collect a dataset across different environments to train the OCR-VN to predict the safety value function, directly from raw observations and uncertainty bounds.
During the deployment phase, we query the OCR-VN with the onboard observation and an estimate of the dynamical uncertainty to construct an adaptive safety filter across different locomotion policies and environments.
The OCR framework is described in more detail next.

\subsection{Training Phase of the OCR Framework}\label{sec:training}

\subsubsection{System Dynamics Model}\label{sec:model}

During the training phase, we aim to distill the ground-truth value functions for a diverse range of environments $e$ into an OCR-VN.
The first difficulty that we encounter is how to model the full quadruped dynamics $f(x, u; e)$, which is high-dimensional and complex.
Our key insight is that many quadruped control schemes are hierarchically composed of a high-level navigation planner and a low-level locomotion policy.
From the perspective of the high-level planner, the quadruped, along with its locomotion policy, form a system with a \textit{reduced-order} dynamics model.
Thus, we propose to use a reduced-order dynamics model $f_r(x_r, \omega)$, where $x_r \in \mathcal{X}_r \subseteq \mathcal{X}$ is the reduced state and $\omega$ is the control input of the reduced-order model.
We capture any possible modeling errors of $f_r(x_r, \omega)$ as disturbances in the system.
We propose to estimate bounds on these disturbances during deployment for adaptive safety guarantees.

We set $f_r(x_r, \omega)$ to the dynamics of a 3D Dubins car system with state $x_r = \left(p_x, p_y, p_\theta\right)$, where $(p_x, p_y)$ is the quadruped's 2D location, and $p_\theta$ is the quadruped's heading.
We model the error in $f_r(x_r, \omega)$ as an unknown additive disturbance $d^e_r$, which is a function of the underlying environment $e$.
Thus, the reduced-order dynamics $f_r(x_r, \omega, d^e_r)$ is given as:
\begin{align} \label{eq:model}
    &&\dot{p_x} = v \cos{p_\theta} + d^e_{p_x}, &&\dot{p_y} = v \sin{p_\theta} + d^e_{p_y}, &&\dot{p_\theta} = w + d^e_{p_\theta}, 
\end{align}
with the control input $\omega = \left(v, w\right)$ being a commanded twist that includes forward velocity $v \in [v_{\min}, v_{\max}]$ and yaw rate $w \in [w_{\min}, w_{\max}]$.
The disturbance input $d^e_r = \left(d^e_{p_x}, d^e_{p_y}, d^e_{p_\theta}\right)$ consists of a bounded additive disturbance in position $|| [d^e_{p_x}, d^e_{p_y}] || \leq \bar{d}^e_{p_{x},p_{y}}$ and a bounded additive disturbance in heading $| d^e_{p_\theta} | \leq \bar{d}^e_{p_\theta}$.
We use $\bar{d}^e_r$ to denote the disturbance bound tuple $\bar{d}^e_r = \left(\bar{d}^e_{p_x,p_y},\bar{d}^e_{p_\theta}\right)$, which is a function of the underlying environment $e$.
Since the disturbances are additive in all state variables, the disturbance bounds can always be chosen large enough to contain any possible modeling errors, although at the cost of model conservatism.
In order to be robust to the modeling error, we assume that the disturbances are adversarial in nature.

Due to the choice of state variables in the reduced-order model, we are restricted to safety constraints specified in terms of the quadruped's 2D location and 1D orientation.
This is sufficient for our work, since we focus exclusively on safe navigation.
We remark that other $f_r$ and disturbance assumptions can be chosen depending on the desired performance, conservatism, and computational complexity.

\subsubsection{Data Generation}\label{sec:data}

Based on our modeling choices in Section \ref{sec:model}, we let the underlying environment $e = \left(\mathcal{F}^e, \bar{d}^e_r\right)$ consist of the failure set $\mathcal{F}^e$ and the disturbance bound $\bar{d}^e_r$ described in Section \ref{sec:model}.
We generate each environment $e$ by randomly spawning 2D obstacles at various locations and sampling a disturbance bound.
These disturbance bounds will correspond to the model uncertainties considered during the deployment phase and will be estimated online (Section \ref{sec:estimator}).

For each of the generated environments, we compute the initial-time ground-truth value function $V(x_r, 0; e)$ using the $\texttt{hj\_reachability}$ Python toolbox \cite{hj_reachability_python}, which we denote as $V(x_r; e)$ for brevity.
We compute the converged value function, that is, when we take the time horizon $T \rightarrow \infty$.
This ground-truth value function is then used to generate data for training the OCR-VN.

Specifically, for each training environment, we generate training pairs $\left( \left( x_r, \bar{d}^e_r, o^e \right), V\left( x_r; e \right) \right)$ by setting the system origin to different states in the environment and rendering the corresponding observation $o^e$ at the origin.
For each system origin and its corresponding observation $o^e$, we sample the value function $V\left( x_r; e \right)$ and its spatial gradients $\nabla V\left( x_r; e \right)$ at uniformly random states $x_r$ (specified in the frame of the new system origin) that are not occluded by obstacles.
The collected data is then used to train the OCR-VN via supervised learning as described in the next section.
Details on the specific parameters of our data generation are provided next.

To generate an environment, we spawn $n$ circular 2D obstacles, where $n$ is uniformly sampled from $\{ 1, ..., 10 \}$.
The radius $r$ of each obstacle is uniformly sampled from $[0.1, 1]$ m.
The location $\left( p_x,p_y \right)$ of each obstacle is uniformly sampled from $[-5, 5]$ m $\times$ $[-5, 5]$ m.
Disturbance bounds $\bar{d}^e_{p_x,p_y}$ and $\bar{d}^e_{p_\theta}$ are uniformly sampled from $[0, 1]$ m/s and $[0, 2]$ rad/s, respectively.
In this work, we use LiDAR observations which consists of $100$ evenly spaced angles from $[-\pi, \pi]$ rad and clipped to within $[0.2, 10]$ m, determined by hardware limits.

For obtaining the ground-truth value function, we use a grid of shape $\left( 100, 100, 60 \right)$ spanning $\left[ -5, 5 \right]$ m $\times$ $\left[ -5, 5 \right]$ m $\times$ $\left[ -\pi, \pi \right]$ rad and a time horizon of $2$ s, when the value function approximately converges.
The control bounds for the system are $v \in \left[ 0, 2 \right]$ m/s and $|w| \leq 2$ rad/s.
We generate a total of $1,000$ training environments and $100$ validation environments.
Generating the datasets takes roughly $35$ minutes on an NVIDIA 3090Ti GPU.
The ground-truth value function $V(x_r; e)$ and the corresponding LiDAR observation $o^e$ for a validation environment are shown in Figure \ref{fig:validation}.

We generate a training batch by first sampling $10$ training environments.
For each environment, we set the system origin to $10$ different states sampled outside of the obstacle set and capture the corresponding LiDAR observation, resulting in $10$ different egocentric observations.
This is done to increase the diversity of LiDAR observations seen during training without substantially increasing the computational effort, which is important because the observations are a high-dimensional network input.
For each sampled system origin, we query the ground-truth value function and its spatial gradients at $500$ random states that are not occluded by obstacles.
This ultimately results in $N = 50,000$ samples per training batch.

\input{figures/validation}

\subsubsection{OCR-VN Architecture and Training}\label{sec:network}

We train an OCR-VN to predict $V(x_r; e)$ from the reduced system state $x_r$, disturbance bound $\bar{d}^e_r$, and the LiDAR observation $o^e$ captured at the system origin.
Let us denote the value predicted by the OCR-VN as $V_{\psi}(x_r, \bar{d}^e_r, o^e)$, where $\psi$ denotes the learnable weights of the neural network.
We choose $V_{\psi}$ to be a multilayer perceptron with $4$ hidden layers of $512$ neurons each.
We desire $V_{\psi}$ to accurately model the spatial gradients of $V$ as well, since they will be used in the safety filter construction.
Thus, we use sinusoidal activations, which have been shown to lead to more effectively gradient modeling than ReLU activations \cite{bansal2021deepreach}.

For each training batch as described in Section \ref{sec:data}, we minimize the MSE training loss:
\begin{align*}
    \frac{1}{N}\sum^{N}_{i=1}(||V^i_{\psi}(x_r, \bar{d}^e_r, o^e) &- V^i(x_r; e)||^2_2 + \\
    ||\nabla_{x} V^i_{\psi}(x_r, \bar{d}^e_r, o^e) &- \nabla_{x} V^i(x_r; e)||^2_2)
\end{align*}
which includes a loss term for both the value and its spatial gradients.
We use the Adam optimizer with a learning rate of $10^{-5}$.
Training converges in roughly $8$ hours on an NVIDIA 3090Ti GPU.
As shown in Figure $\ref{fig:validation}$, the OCR-VN can accurately model the value function and its spatial gradients for a validation environment using a single LiDAR observation $o^e$.

\subsubsection{OCR-VN Calibration} \label{sec:calibration}

The OCR-VN will inevitably contain learning errors that can be critical for safety.
To safeguard against these errors, we calibrate the OCR-VN by shifting its output by a probabilistic error bound $\delta$ computed on the validation dataset.
We will refer to $\delta$ as the ``calibration level'' and the shifted network output as the ``calibrated output''.
We employ conformal prediction, a popular uncertainty quantification tool in the machine learning literature, to compute $\delta$ up to a desired confidence $\beta$ and violation rate $\epsilon$ \cite{MAL-101,pmlr-v242-lin24a}.

Our procedure is as follows.
Select a desired confidence $\beta \in (0, 1)$ and violation rate $\epsilon \in (0, 1)$.
Sample $N$ calibration points according to a distribution $\mathbb{P}$ over the validation dataset $\Delta$.
Compute the conformal scores $\{s_i\}_{i=1}^N$ as the prediction errors $s_i = V^i_\psi \left( x_r, \bar{d}_r^e, o^e \right) - V^i\left( x_r; e \right)$.
The following theorem provides a probabilistic bound on the OCR-VN error:
\newtheorem{theorem}{Theorem}
\begin{theorem}[Conformal OCR-VN Calibration] \label{thm:calibration}
    Compute the number of ``outliers'' $k$ as:
\vspace{-0.5em}
    \begin{equation} \label{eq:calibration_relationship}
        \begin{aligned}
            \argmax_k k: \sum^k_{i=0} \binom{N}{i} \epsilon^i (1-\epsilon)^{N-i} \le \beta,
        \end{aligned}
    \end{equation}
    where $\beta$, $\epsilon$, and $N$ are as defined above.
    Compute the calibration level $\delta$ as the $\frac{N-k}{N}$ quantile of the conformal scores $\{s_i\}_{i=1}^N$.
    Then, with probability at least $1-\beta$ over the draws of the calibration samples, the following holds:
    \begin{equation} \label{eq:calibration_guarantee}
        \begin{aligned}
            \underset{(x_r, e, \bar{d}^e_r, o^e) \in \Delta}{\mathbb{P}} \left( V_{\psi}(x_r, \bar{d}^e_r, o^e) - V(x_r; e) > \delta \right) \le \epsilon
        \end{aligned}
    \end{equation}
\end{theorem}
The proof of Theorem \ref{thm:calibration} is presented in Appendix \ref{apd:proof}.
Ignoring the confidence parameter $\beta$ for a moment, Inequality $\eqref{eq:calibration_guarantee}$ tells us that the volume of the validation dataset for which the OCR-VN overestimates the safety value by more than $\delta$ is bounded by the violation parameter $\epsilon$.
This enables us to be sure, up to a desired $\epsilon$, that the true safety value is at least as large as that predicted by the OCR-VN, once adjusted by $\delta$.

To interpret the confidence parameter $\beta$, note that the calibration level $\delta$ is a random variable that depends on the randomly sampled calibration set.
It might be the case that we happen to draw an unrepresentative calibration set, in which case the $\epsilon$ bound does not hold.
$\beta$ controls the probability of this adverse event, which regards the correctness of the probabilistic guarantee given by Inequality $\eqref{eq:calibration_guarantee}$.
Fortunately, $\beta$ goes to $0$ exponentially with $N$, so $\beta$ can be chosen sufficiently small, such as $10^{-12}$, when we sample large $N$.
$1-\beta$ will then be so close to $1$ that it does not have any practical importance.

We set $N=10^7, \beta=10^{-12}$ and compute $\delta$ corresponding to various $\epsilon$ in Table \ref{tab:calibration}.
We select $\delta = 0.49$ m associated with $\epsilon = 10^{-2}$ as a comfortable trade-off between conservatism and performance for the tasks in this work.

\input{tables/calibration_table}

\subsection{Deployment Phase of the OCR Framework}\label{sec:deployment}

In this section, we describe the steps involved in deploying the OCR framework onto a quadruped robot.
During deployment, the quadruped periodically receives a LiDAR observation $o^e$ from an onboard sensor as a function of the environment $e$.
Using a LiDAR-based state localization algorithm, the quadruped maintains an estimate of its reduced state $x_r$ in the relative frame defined by $o^e$.
The quadruped then estimates the disturbance bound $\bar{d}^e_r$ as a function of the most recent state and action history $\left( x^{i-k:i}_r, \omega^{i-k:i-1} \right)$ using an online disturbance bound estimation scheme.
It queries the OCR-VN with the input $\left( x_r, \bar{d}^e_r, o^e \right)$ to construct an adaptive safety filter that minimally overrides the nominal controller when necessary to maintain safety.
We describe the details of the online disturbance bound estimation scheme and the adaptive safety filter next.

\subsubsection{Online Disturbance Bound Estimation}\label{sec:estimator}

We propose to estimate the disturbance bound $\bar{d}^e_r$ using the most recent state and action history, to enable rapid adaptation to dynamical uncertainty.
Let $x^{i-k:i}_r$ and $\omega^{i-k:i-1}$ denote a discrete history of estimated states and twist commands ending at time $\tau$.
To efficiently roll out the state trajectory in discrete time for online computation, we use Euler's method.
Since disturbance is additive in $f_r(x_r, \omega, d^e_r)$ in Equation \eqref{eq:model}, we have:
\begin{align*}
    x^i_r &\approx x^{i-k}_r + \sum_{j=i-k}^{i-1} \eta \cdot ( f_r(x^j_r, \omega^j) + \left(d^e_r\right)^j ) \\
    &\approx x^{i-k}_r + \sum_{j=i-k}^{i-1} \eta \cdot f_r(x^j_r, \omega^j) + \eta \sum_{j=i-k}^{i-1} \left(d^e_r\right)^j, \\
\end{align*}
where $\eta$ is the discrete time step.
To ensure that we capture systemic disturbances in the dynamics and ignore transient noise in the state estimation, we assume that the disturbance input takes the form of a low-frequency signal and remains a constant $d^\tau_r$ throughout the short history:
\begin{align*}
    x^i_r &\approx x^{i-k}_r + \sum_{j=i-k}^{i-1} \eta \cdot f_r(x^j_r, \omega^j) + \eta \cdot k \cdot d^\tau_r. \\
\end{align*}
Thus, we estimate the disturbance $d^\tau_r$ as:
\begin{equation}
    d^\tau_r \approx \frac{x^i_r - \hat{x}^i_r}{\eta \cdot k},
\end{equation}
where $\hat{x}^i_r \coloneq x^{i-k}_r + \sum_{j=i-k}^{i-1} \eta \cdot f_r(x^j_r, \omega^j)$ is the state predicted by disturbance-free dynamics.
Intuitively, $d^\tau_r$ is the error in the state prediction normalized by the prediction time horizon $\eta \cdot k$.
We set $\eta \cdot k = 2$ s in practice.

By computing $d^\tau_r$ over the most recent $\varphi$ discrete time steps ending at the current time $t$, i.e., for $\tau = t - j \cdot \eta, \forall j \in \left\{ 0, 1, ..., \varphi - 1 \right\}$, we construct a sliding window of estimated disturbances $d^{1:\varphi}_r$.
To ensure good coverage of the estimated disturbances, we compute the disturbance bound $\bar{d}^e_r$ as:
\begin{equation} \label{eq:dst}
    \bar{d}^e_r = | \mu( \left(d^{1:\varphi}_r\right)_c ) | \pm b \cdot \sigma( \left(d^{1:\varphi}_r\right)_c ),
\end{equation}
where the coverage parameter $c \in [0, 1]$ determines the middle fraction of the sorted window of disturbances $d^{1:\varphi}_r$ to be considered.
This makes sure that the safety value function does not become overly conservative due to outlier disturbances.
$\mu (\cdot)$ is the sample mean, and $\sigma (\cdot)$ is the sample standard deviation.
$b \geq 0$ is the spread of disturbances, in standard deviation units, that we compute the disturbance bound for, modulating the degree of robustness of the disturbance bound.
A larger choice of $b$ results in a more robust disturbance bound.
We set $c=0.8$, $b=2$, and $\eta \cdot \varphi = 2$ s in practice.
We compute two separate disturbance bounds: $\bar{d}^e_{p_x, p_y}$ bounding the uncertainty in the position dynamics and $\bar{d}^e_{p_\theta}$ bounding the uncertainty in the yaw dynamics.
For computing a bound on the norm of $d^e_{p_x, p_y}$, we use Equation \eqref{eq:dst} on a window of disturbance norms instead of raw values.

\subsubsection{Adaptive Safety Filter} \label{sec:filter}

We use the OCR-VN to construct a smooth least-restrictive safety filter (see Section \ref{sec:slr}) safeguarding a potentially unsafe nominal controller that is given to us.
Specifically, we compute a filtered high-level policy $\pi_\text{safe}^\text{high}$ which, when combined with the underlying low-level policy $\pi_\text{nom}^\text{low}$, maintains the overall system safety.

We compute the hierarchical filtered policy $\pi_\text{safe}(x, \bar{d}^e_r, o^e)$ as follows:
\begin{equation} \label{eq:filter}
    \pi_\text{safe}(x, \bar{d}^e_r, o^e) = 
    \begin{cases}
        \pi_\text{nom}(x), & V_{\psi}(x_r, \bar{d}^e_r, o^e) > \delta \\
        \pi_\text{nom}^\text{low} \circ \pi_\text{QP}^\text{high}(x_r, \bar{d}^e_r, o^e), & V_{\psi}(x_r, \bar{d}^e_r, o^e) \leq \delta,
    \end{cases}
\end{equation}
where $\pi_\text{QP}^\text{high}(x_r, \bar{d}^e_r, o^e)$ is obtained by solving:
\begin{equation}
    \begin{aligned} \label{eq:qp}
        &\min_{\omega \in \Omega, s \ge 0} \left( || \omega - \pi_\text{nom}^\text{high}(x_r) ||_2^2 + \lambda s^2 \right) \\
        &\text{s.t.} \min_{d_r \in \mathcal{D}^e_r} \nabla V_{\psi}(x_r, \bar{d}^e_r, o^e) \cdot f_r(x_r, \omega, d_r) \ge -s
    \end{aligned}
\end{equation}
where the slack variable $s$ ensures that there always exists a feasible solution, and $\lambda$ (set to $10^3$) is the relative weight of $s$ in the objective.
$\mathcal{D}^e_r$ is the set of disturbances respecting the disturbance bound $\bar{d}^e_r$.
Intuitively, \eqref{eq:qp} minimally adjusts the nominal policy so that the resultant twist commands ensure system safety in the current environment.
Furthermore, the introduction of the slack variable ensures that the QP problem in \eqref{eq:qp} is always feasible, especially when the disturbance bounds are overly conservative.
Note that the disturbance optimization in the constraint in \eqref{eq:qp} is independent of the optimization of $\omega$, since the disturbance enters additively into $f_r$ in Equation \eqref{eq:model}.
Since $f_r$ is affine in control and disturbance, \eqref{eq:qp} is a quadratic program (QP) that we can solve efficiently online, making the proposed framework amenable for real-world robotic systems.

We remark that the design of the filter in \eqref{eq:filter} is grounded in reachability theory.
If $V_{\psi}$ perfectly models the underlying value function and the system starts outside of the $\text{BRT}$, then $\eqref{eq:filter}$ is guaranteed to maintain safety \cite{10665911}.

\begin{remark}
    When the quadruped is controlled by an \textit{end-to-end} nominal controller, we can use the OCR-VN framework to maintain safety with a minor adjustment.
    While $V_\psi > 0$, we permit the nominal controller to execute unhindered.
    Otherwise, we must provide our own high-level planner $\pi_\text{nom}^\text{high}(x)$ for solving \eqref{eq:qp} and a low-level policy $\pi^\text{low}_\text{nom}$ for execution.
\end{remark}

%% file: figures/framework.tex
\setcounter{figure}{1}
\begin{figure*}[!t]
\centering
\includegraphics[width=1.0\textwidth]{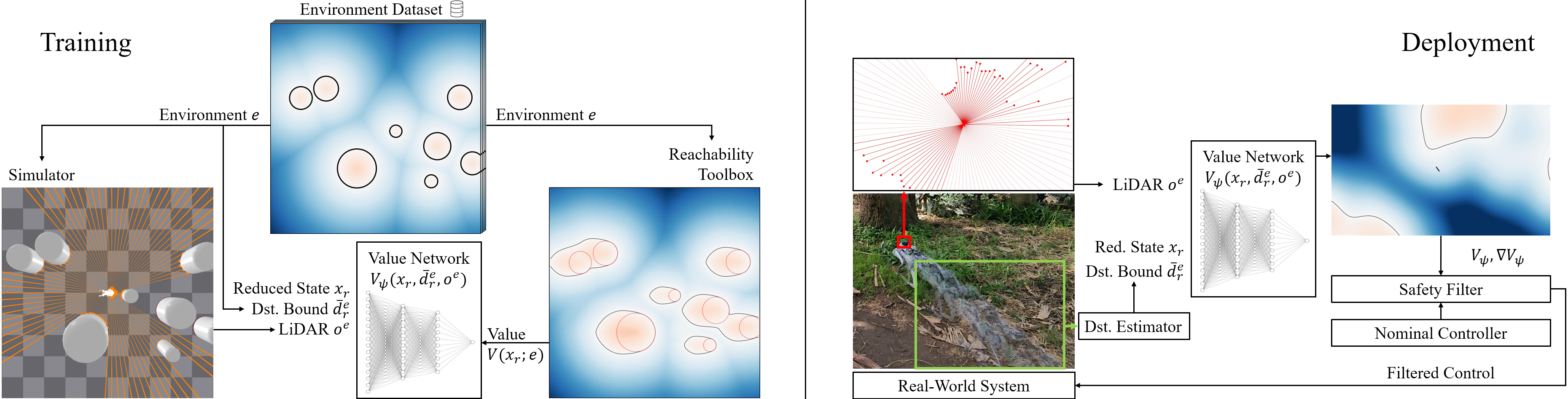}
\caption{The OCR framework. (Left) During training, we generate environments with random obstacles and disturbance bounds. The OCR-VN is trained to predict the value function (visualized over a grid) using the disturbance bound, the LiDAR reading, and the state. (Right) During deployment, the OCR-VN is queried with the observed LiDAR reading, the disturbance bound estimated using the most recent state and action history, and the current state estimate to construct an adaptive safety filter.}
\label{fig:framework}
\end{figure*}

%% file: figures/validation.tex
\begin{figure}[!t]
\centering
\includegraphics[width=0.495\textwidth]{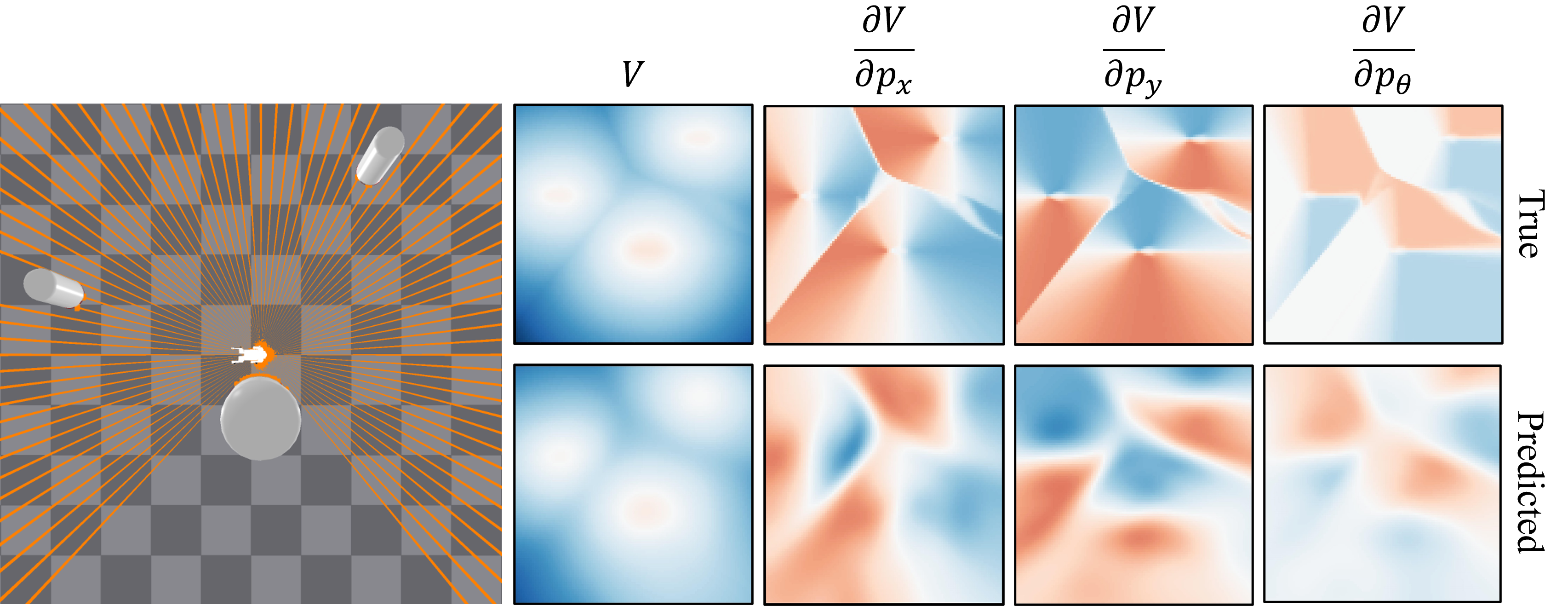}
\caption{(Left) A LiDAR observation $o^e$ in a validation environment where $\bar{d}^e_{p_{x},p_{y}}=0.82$ m/s, $\bar{d}^e_{p_\theta}=0.56$ rad/s. (Right top-row) The ground-truth value function and its spatial gradients. (Right bottom-row) OCR-VN predictions using $o^e$ and $\bar{d}^e_r$. As shown above, the OCR-VN predictions for the value function and its spatial gradients are highly accurate.}
\label{fig:validation}
\end{figure}

%% file: tables/calibration_table.tex
\begin{table}[!t]
\caption{Calibration Levels for the OCR-VN \\
with $N = 10^7, \beta = 10^{-12}$\label{tab:calibration}}
\centering
\begin{tabular*}{\linewidth}{@{\extracolsep{\fill}} l@{\qquad}|c@{\qquad}|c@{\qquad}|c@{\qquad} }
\toprule
$\textbf{Violation Rate }\bm{\epsilon}$ & $10^{-1}$ & $10^{-2}$ & $10^{-3}$ \\
$\textbf{Calibration Level }\bm{\delta}\textbf{ (m)}$ & $0.22$ & $0.49$ & $0.99$ \\
\bottomrule
\end{tabular*}
\end{table}

%% file: sections/simulation_experiments.tex
\section{Simulation Experiments}\label{sec:simulation_experiments}

We test the proposed framework on a Unitree Go1 quadruped in Isaac Sim (this section) and on a hardware testbed (Section \ref{sec:hardware_experiments}) in a variety of unseen obstacle settings and terrains, as well as on different nominal controllers.
The goal of our simulation studies and experiments is to answer the following questions about the proposed framework:
\begin{enumerate}
    \item Can the safety filter dynamically adapt to unknown obstacles to maintain safety?
    \item Can the safety filter adapt to system and environment uncertainty (e.g., uneven terrains, slippery surfaces, etc.)?
    \item Can the safety filter maintain safety under different nominal policies?
\end{enumerate}

\subsection{Nominal Controllers}\label{sec:controllers}

We test our framework on several different hierarchical nominal policies.
A nominal policy is generated by selecting one of the three high-level planners and one of the three low-level locomotion policies described below.
These policies are selected to contain both RL-based policies as well as MPC-based policies.
We test all different combinations of high-level and low-level policies, but present the results for only a few of them here for brevity purposes.

Additionally, we include an end-to-end nominal policy from ABS \cite{he2024agile} (called ABS-Agile here on), as an illustration of how our method can generalize to end-to-end nominal policies.
To implement our method on top of the end-to-end ABS-Agile policy, we compute the safety twist commands using the proposed filter framework in Equation \eqref{eq:filter} with the Predictive Sampling-based (PS) high-level planner (see \ref{sec:PS}) and use the ABS-Recovery policy (see \ref{sec:ABS-Recovery}) to track the filtered twist commands.
Whenever the system safety is not at risk, we use the nominal ABS-Agile policy.

\subsubsection{High-Level Planners}

\paragraph{Predictive Sampling-Based Planner (PS)}\label{sec:PS}

PS is a sampling-based high-level MPC planner that optimizes the robot trajectory to reach the goal while avoiding obstacles.
The cost function for a discrete control sequence $\bm{\omega} = \{\omega^j\}_{j=1}^{\lfloor T/\eta \rfloor}$ evaluated at the reduced state $x_r$ is given by:
\begin{align*}
    J_{\bm{\omega}}(x_r) = \sum_{j=1}^{\lfloor T/\eta \rfloor} \left( ||\hat{x}^j_r - g|| + c\cdot \mathbbm{1}\{\hat{x}^j_r \in \hat{\mathcal{L}}^e\} \right)
\end{align*}
where $T$ is the prediction horizon and $\eta$ is the discrete time step.
$\hat{x}^j_r$ is the future state at time $j \cdot \eta$ predicted by the reduced-order dynamics $f_r$ in Equation \eqref{eq:model} after applying $\bm{\omega}$ to the system.
$g$ is the goal state, $c$ is a collision penalty, and $\hat{\mathcal{L}}^e$ is the estimated obstacle map computed online using obtained LiDAR scans via $\texttt{tinySLAM}$ \cite{5707402}.
Intuitively, the cost function is minimized by control sequences that bring the system to the goal fastest without collisions.
To compute the optimal high-level commands, PS samples $N$ control sequences $\bm{\omega}_i, \forall i = 1, ..., N$ from a Gaussian centered at $\bm{\omega}_\text{seed}$ with standard deviation $\sigma$.
The cost $J_{\bm{\omega}_i}(x_r)$ for each sequence is computed and finally the best control sequence $\bm{\omega}_\text{best}$ is selected which minimizes the cost.
PS executes the first twist command in $\bm{\omega}_\text{best}$ and sets $\bm{\omega}_\text{seed} = \bm{\omega}_\text{best}$ for the next control iteration.
We set $N=1,000$, $\sigma=0.5$, $T=4$ s, $\eta=0.2$ s, $c=10^9$, and the initial $\bm{\omega}_\text{seed}$ to the control range center.

\paragraph{Naive Planner (NVE)}

NVE performs basic goal-seeking without obstacle avoidance.
Let $p_{\theta_\text{goal}}$ be the angle to the goal relative to the robot's heading.
NVE computes:
\begin{align*}
    v &= \begin{cases}
        v_{\max}, & |p_{\theta_\text{goal}}| \le \frac{\pi}{2} \\
        v_{\min}, & |p_{\theta_\text{goal}}| > \frac{\pi}{2},
    \end{cases} \\
    w &= \begin{cases}
        w_{\min} \cdot \min \{ | \frac{p_{\theta_\text{goal}}}{p_{\theta_{\max}}} |, 1\}, & p_{\theta_\text{goal}} \le 0 \\
        w_{\max} \cdot \min \{ | \frac{p_{\theta_\text{goal}}}{p_{\theta_{\max}}} |, 1\}, & p_{\theta_\text{goal}} > 0,
    \end{cases}
\end{align*}
where $p_{\theta_{\max}}$ is the heading difference beyond which yaw rate is maximized.
We set $p_{\theta_{\max}}=\frac{\pi}{4}$ rad.
Intuitively, NVE turns greedily towards the goal and commands maximum/minimum forward velocity when the goal is ahead/behind.

\paragraph{Human Teleoperation (HMN)}

a human teleoperator provides the high-level twist commands.

\subsubsection{Low-Level Policies}

\paragraph{Walk-These-Ways (WTW)}

an RL-based policy that encodes a structured family of locomotion strategies to enable diverse task generalization such as crouching, hopping, running, and others \cite{margolis2022walktheseways}.
During training, the policy is rewarded for accurate twist tracking and gait following.
During deployment, different locomotion strategies can be used by changing gait parameters.
We use the gait parameters corresponding to a 3 Hz trot for its stability and agility.

\paragraph{Legged Model Predictive Control (MPC) by Unitree}

a model-based policy provided by Unitree Robotics as a ROS package for the Go1 quadruped robot \cite{unitree_ros}.
The policy tracks twist commands using an internal MPC-based algorithm that computes motor torque controls.
It is widely used by consumers of the Go1 quadruped hardware.

\paragraph{ABS Recovery Policy}\label{sec:ABS-Recovery}

an RL-based policy used in the ABS framework to track twist commands as fast as possible in order to serve as a backup shielding policy \cite{he2024agile}.
During training, the policy is rewarded for accurate twist tracking and maintaining a stable posture.
Similar to its role in the ABS framework, we will use the ABS-Recovery policy as the low-level backup shielding policy for the ABS-Agile policy. 

\subsection{Baseline and Ablations}

\input{figures/splash}

We evaluate our framework against a range of nominal controllers and the end-to-end safety framework, ABS, proposed in \cite{he2024agile}.
ABS integrates an RL-based agile policy with a safety-oriented recovery policy.
A value function-based predictor determines when to switch to the recovery policy, which then generates safe twist commands to be executed by the recovery policy.
We refer the interested readers to \cite{he2024agile} for more details.

We also evaluate the OCR framework without the disturbance estimation module (OCR $\backslash$ DE) and calibration step (OCR $\backslash$ C), as well as if the safety filter is not used at all (No Filter) to understand the importance of each of these modules.

\subsection{Metrics}

Each experiment trial ends in either a success, a collision, or a timeout after $60$ s.
Across successful runs, we report the average velocity $\bar{v}$ of the quadruped along the trajectory, the average rate of safety filter activation $\bar{r}$, and the average minimum distance to the obstacle set $\bar{q}$.

\subsection{Simulation Setups}

We test our framework as well as baselines on randomly generated environments.
Each environment contains $4$ circular obstacles, where each obstacle is spawned at some location drawn uniformly randomly from $[-2, 2]$ m $\times$ $[-2, 2]$ m.
Each obstacle has a radius drawn uniformly randomly from $[0.1, 1]$ m.
Each environment also draws an additional payload uniformly randomly from $\left[-1, -0.5\right] \cup \left[0.5, 1\right]$ kg and a ground friction coefficient $\left[0.5, 0.75\right] \cup \left[1.25, 1.5\right]$ for dynamical variation.
We remark that these payloads and frictions are contained within the range used to train the nominal policies and baselines so that the comparison is fair, but they represent the outer limits to effectively test robustness.
The quadruped must navigate from $(-5, 0)$ m to $(5, 0)$ m without colliding (see Figure \ref{fig:splash}).

Additionally, we present hand-designed environments that stress-test and highlight the abilities of our framework.

\subsection{Simulation Results}

\input{tables/simulation_results_table}

Recall that the goal of our simulation studies and experiments is to answer the following questions: can the safety filter adapt to unknown obstacles in the environment, uncertainty in the system and environment, and different nominal policies?

\subsubsection{Safeguarding Different Nominal Controllers}

We first evaluate the efficacy of the OCR framework and its ablations for safeguarding various nominal controllers in different obstacle settings.
Numerical results are listed in Table \ref{tab:simulation_results}.
The OCR framework achieves high success rates, low collision rates, and comfortable distances to the obstacle set across all nominal controllers, highlighting its ability to automatically safeguard different nominal controllers in different environments without \textit{a priori} knowledge.
Its success rates are consistently high regardless of whether the underlying nominal control has built-in obstacle avoidance (ABS-Agile and PS + WTW) or not (NVE + WTW).
Figure \ref{fig:splash} illustrates the robot trajectories under the OCR framework and the framework's ability to ensure safety when the nominal controller would cause collision.

The results in Table \ref{tab:simulation_results} also clearly show the importance of the disturbance estimation module and the calibration step for enabling the OCR framework to be robust to modeling and learning errors, respectively.
Interestingly, even though the ABS framework is particularly designed to safeguard the ABS-Agile policy, our method also outperforms the ABS framework on ABS-Agile.
A key reason for this is the robustness of the proposed framework not only to environment uncertainty (i.e., unknown obstacles) but also to dynamics uncertainty, as we discuss next.

\subsubsection{Robustness to Dynamical Uncertainty}

Results in Table \ref{tab:simulation_results} demonstrate the ability of the OCR framework to handle variation in the system dynamics.
Recall that each environment has its own payload and friction parameters, which significantly influence the dynamics of the system.
Additionally, different low-level policies have different abilities to track velocity and yaw rate commands.
Despite these challenges, the OCR framework consistently achieves a high success rate without a priori access to the underlying nominal policy or simulation environments.

\input{figures/safety_rates}

\input{figures/disturbance_estimation}

To study robustness more thoroughly, we plot safety rates across environment settings for different frameworks using the ABS-Agile policy as the nominal policy in Figure \ref{fig:safety_rates}.
The OCR framework achieves high safety rates across all settings, surpassing the ABS baseline particularly in low-friction environments, but only when using disturbance estimation.
This leads us to conclude that disturbance estimation plays a key role in the robustness of the proposed framework.
For all numerical results, see Appendix \ref{apd:results}.

\input{figures/sim_combined}

Figure \ref{fig:disturbance_estimation} (a) visualizes the behavior of the OCR and ABS frameworks in an environment with high perturbations in dynamics (a payload of $-0.9$ kg and a friction of $0.5$).
The OCR framework estimates large disturbances online and accordingly intervenes to steer the system away from collision.
The adaptation facilitated by the disturbance estimation module is visualized in Figure \ref{fig:disturbance_estimation} (b) in a hand-designed environment with a sudden friction change in the blue region.
Correspondingly, the proposed framework automatically estimates larger disturbance bounds in this region of low friction (the purple point), which subsequently, leads to a more aggressive safety filter to ensure system safety.
The ABS framework, in contrast, intervenes later and allows the quadruped to approach dangerously close to the wall obstacle, eventually leading to collision.

\subsubsection{Further Comparisons with ABS}

We now further compare the proposed OCR framework with ABS, as it leverages a similar safety filtering framework.
One of the key advantages of the OCR framework is that it grants the user the freedom to choose the most appropriate nominal controller for the task at hand without needing to recompute safety assurances, whereas the ABS framework is specific to a particular nominal policy.
We demonstrate this utility in Figure \ref{fig:sim_combined} (a), where the robot must navigate two sharp turns.
The OCR framework successfully completes the task using the PS + WTW nominal controller.
On the other, the ABS baseline gets stuck, because the learning-based ABS-Agile policy is not suited for the required navigation style.

The policy-independence of the OCR framework is a consequence of the \textit{optimality} of the underlying value function synthesized during the training phase described in Section \ref{sec:training}.
We demonstrate the advantages of using the optimal value function in Figure \ref{fig:sim_combined} (b) in a hand-designed ``dead-end'' environment.
Since the setting is highly cluttered, we use OCR $\backslash$ C to reduce conservatism.
We remark that this remains a fair evaluation with respect to the ABS baseline because OCR $\backslash$ C retains a comparable safety rate to ABS in Table \ref{tab:simulation_results}.

Using a control search guided by a policy-conditioned value function, the ABS baseline suffers from unnecessary interventions and suboptimality related to the quality of the ABS-Agile policy.
On the other hand, OCR framework reasons about the optimal safety behavior (regardless of the nominal policy), reducing unnecessary interventions and permitting task progress.

%% file: figures/splash.tex
\begin{figure*}[!b]
\centering
\includegraphics[width=1.0\textwidth]{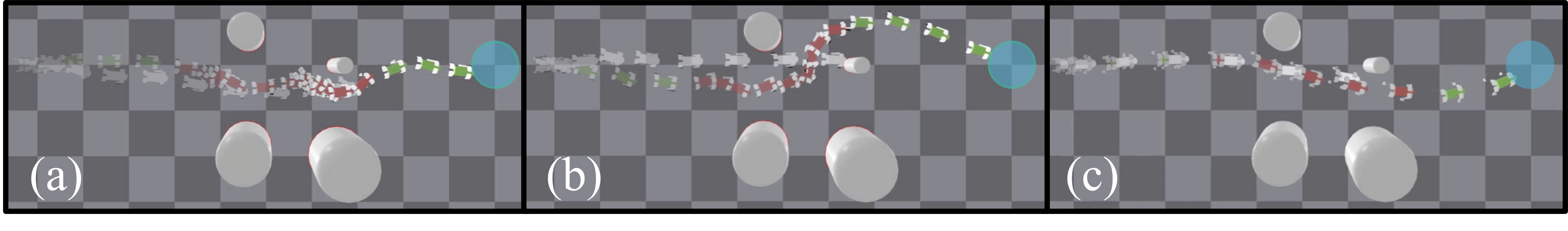}
\caption{The OCR framework ({\color{ForestGreen}green}/{\color{BrickRed}red} : nominal/filtered) safeguards different nominal controllers navigating to a goal ({\color{cyan}cyan}) in an environment with a payload of $-0.6$ kg and a friction of $0.7$. By themselves (white), the (a) PS + WTW, (b) NVE + WTW, and (c) ABS-Agile controllers fail to maintain safety due to dynamical uncertainty caused by low payload and friction.}
\label{fig:splash}
\end{figure*}

%% file: tables/simulation_results_table.tex
\begin{table*}[!t]
\caption{Simulation Results for OCR, Ablations, and ABS across Nominal Controllers (100 Trials)\label{tab:simulation_results}}
\centering
\begin{tabular*}{\linewidth}{@{\extracolsep{\fill}} ll@{\qquad\qquad}|cccccc }

\toprule
\textbf{Controller} & \textbf{Filter} & \textbf{Success Rate $\uparrow$} & \textbf{Collision Rate $\downarrow$} & \textbf{Timeout Rate $\downarrow$} & $\mathbf{\bar{v}}$ \textbf{(m/s) $\uparrow$} & $\mathbf{\bar{r} \downarrow}$ & $\mathbf{\bar{q}}$ \textbf{(m) $\uparrow$} \\
\midrule
    \multirow{5}{*}{ABS-Agile}
      & No Filter                               & 0.75 & 0.25 & 0.00 & \textbf{2.10} & \textbf{0.00} & 0.41 \\
      & ABS & 0.80 & 0.20 & 0.00 & 2.03 & 0.03 & 0.41 \\
      & OCR $\backslash$ DE & 0.40 & 0.35 & 0.25 & 0.98 & 0.62 & 0.41 \\
      & OCR $\backslash$ C            & 0.81 & 0.19 & 0.00 & 1.70 & 0.28 & 0.41 \\
      & \textbf{OCR (ours)}                     & \textbf{0.91} & \textbf{0.09} & 0.00 & 1.22 & 0.59 & \textbf{0.58} \\
\midrule
    \multirow{4}{*}{PS + WTW}
      & No Filter                               & 0.72 & 0.28 & 0.00 & \textbf{1.36} & \textbf{0.00} & 0.43 \\
      & OCR $\backslash$ DE & 0.78 & 0.22 & 0.00 & 0.84 & 0.61 & 0.42 \\
      & OCR $\backslash$ C            & 0.90 & 0.09 & 0.01 & 1.21 & 0.18 & 0.44 \\
      & \textbf{OCR (ours)}                     & \textbf{1.00} & \textbf{0.00} & 0.00 & 0.97 & 0.42 & \textbf{0.66} \\
\midrule
    \multirow{4}{*}{NVE + WTW}
      & No Filter                               & 0.20 & 0.80 & 0.00 & \textbf{1.70} & \textbf{0.00} & 0.31 \\
      & OCR $\backslash$ DE & 0.21 & 0.79 & 0.00 & 0.93 & 0.61 & 0.37 \\
      & OCR $\backslash$ C            & 0.45 & 0.55 & 0.00 & 1.32 & 0.28 & 0.39 \\
      & \textbf{OCR (ours)}                     & \textbf{0.91} & \textbf{0.08} & 0.01 & 1.04 & 0.47 & \textbf{0.62} \\
\bottomrule

\end{tabular*}
\vspace{-0.25cm}
\end{table*}

%% file: figures/safety_rates.tex
\begin{figure}[!t]
\centering
\includegraphics[width=3.5in]{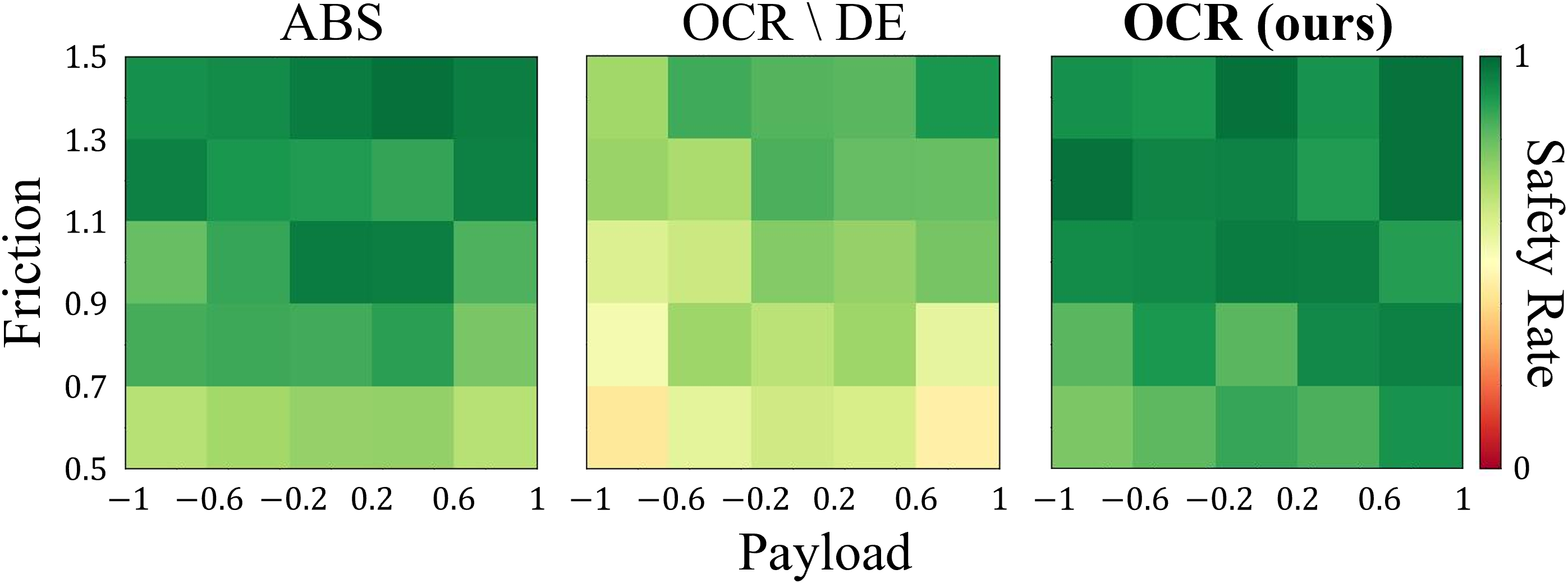}
\caption{(In color) Safety rates across settings (1,000 trials).}
\label{fig:safety_rates}
\vspace{-0.5cm}
\end{figure}

%% file: figures/disturbance_estimation.tex
\begin{figure*}[!b]
\centering
\includegraphics[width=1.0\textwidth]{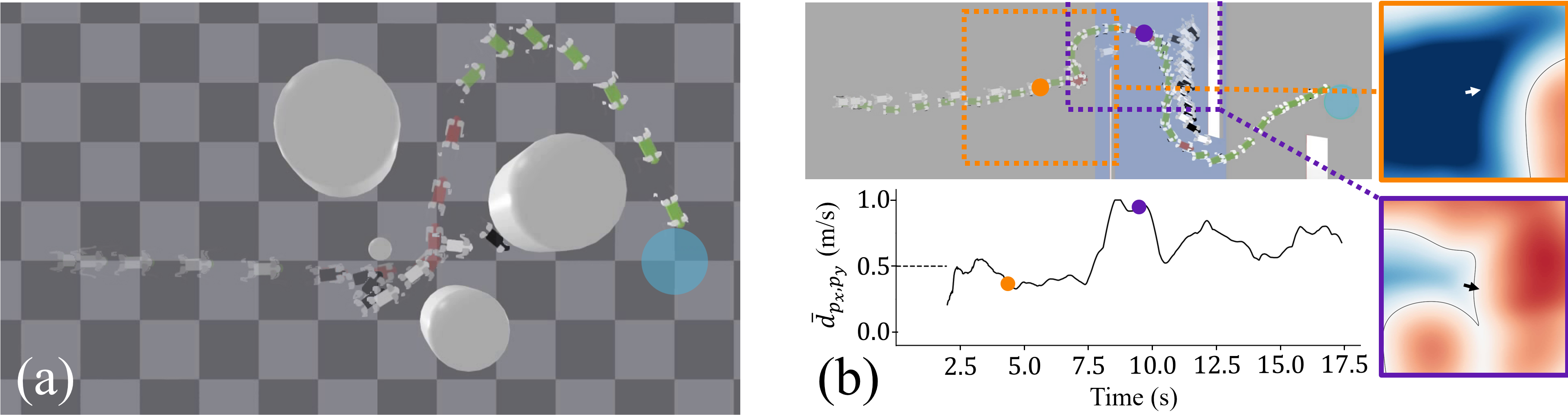}
\caption{OCR ({\color{ForestGreen}green}/{\color{BrickRed}red} : nominal/filtered) and ABS (white/black : nominal/filtered) frameworks in (a) a validation environment with a payload of $-0.9$ kg and a friction of $0.5$ and (b) a hand-designed obstacle configuration with a region of low friction ({\color{blue}blue}). In (b), we plot the evolution of the estimated disturbance bound in position $\bar{d}^e_{p_x,p_y}$, as well as the OCR-VN predictions at two different states ({\color{orange}orange}, {\color{violet}purple}).}
\label{fig:disturbance_estimation}
\end{figure*}

%% file: figures/sim_combined.tex
\begin{figure*}[!t]
\centering
\includegraphics[width=1.0\textwidth]{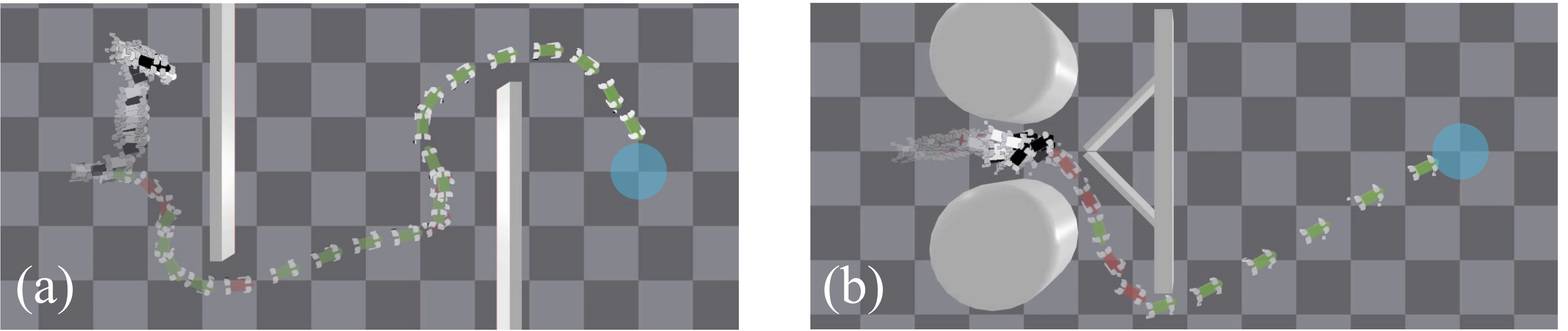}
\caption{OCR ({\color{ForestGreen}green}/{\color{BrickRed}red} : nominal/filtered) and ABS (white/black : nominal/filtered) frameworks in environments with hand-designed obstacle configurations. In (b), the OCR framework is used without calibration due to the highly cluttered setting.}
\label{fig:sim_combined}
\end{figure*}

%% file: sections/hardware_experiments.tex
\section{Hardware Experiments}\label{sec:hardware_experiments}

We deploy the OCR framework with nominal controllers from Section \ref{sec:controllers} on a Unitree Go1 quadruped robot equipped with an onboard Slamtec RPLiDAR A2 sensor.
The quadruped maintains an estimate of its state in the global frame via $\texttt{tinySLAM}$ \cite{5707402}, a LiDAR-based simultaneous localization and mapping (SLAM) algorithm, which is subsequently used for disturbance bound estimation.

\subsection{Hardware Setups}

We first quantitatively and qualitatively compare the proposed framework and all baselines in two different experiment settings.
In the first setting, the robot needs to navigate through an obstacle maze consisting of walls and circular obstacles.
The second setting is the same as the first, except that a rectangular region directly preceding the first obstacle has a very low friction due to an oil-soaked tarp, to test the dynamic safety adaptivity of different methods.
The setups are shown in Figure \ref{fig:slippery}.
In our experiments, we also include results on a few additional nominal policies to stress-test our system.

Additionally, we qualitatively test our framework in a diverse set of real-world experiments, including cluttered environments, rough terrains, external disturbances, and adversarial human teleoperation.

\subsection{Hardware Results}

The hardware results listed in Table \ref{tab:hardware_results} reaffirm the findings in simulation.
Namely, the OCR framework is effective at safeguarding a diverse set of controllers.
The framework also displays a significant robustness to changes in the system dynamics, as evidenced by the high success rates even in the slippery condition, where the ABS baseline fails.
Figure \ref{fig:slippery} illustrates the difference between the OCR and ABS frameworks in the slippery condition. Whereas the ABS baseline intervenes too late to maintain safety, the OCR framework slows down the robot and turns in time to avoid collision.

\input{tables/hardware_results_table}

\input{figures/slippery}

We further demonstrate the capabilities of the framework in a variety of real-world scenarios.
Videos of these demonstrations can be found on the project website listed above Figure \ref{fig:overview}.
Many of these scenarios are illustrated in Figure \ref{fig:overview}.

\subsubsection{Safeguarding Different Nominal Controllers}

Figure \ref{fig:overview} illustrates the ability of the OCR framework to automatically safeguard a variety of high-level planners, including (a) learning-based, (c, f, k) model-based, (b, d, g, h, i, j) human teleoperated, and (e) blind planners, on top of different low-level locomotion policies, including (a, f, i, j, k) learning-based and (b, c, d, e, g, h) model-based policies.

\subsubsection{External Disturbances}

Subplot (f) in Figure \ref{fig:overview} demonstrates the framework's ability to preserve safety even under forceful disturbances.
After receiving a kick while on a slippery floor, the framework guides the robot to turn sharply from the obstacle and move away to maintain safety.

\subsubsection{Adversarial Human Teleoperation}

In subplot (h) of Figure \ref{fig:overview}, an adversarial human teleoperator attempts to collide the robot into a pillar multiple times.
The filter activates only when necessary and causes the robot to veer to avoid collision while respecting the commanded input as much as possible.

\subsubsection{Dynamic Obstacles}

The framework's ability to react to dynamic obstacles in real time is illustrated in subplots (e, k) in Figure \ref{fig:overview}.
The robot slows down from a velocity of roughly $2$ m/s to provide enough turn radius to avoid the box placed suddenly by a human in front of the robot in subplot (e).

\subsubsection{Cluttered Indoor Environments}

The robot navigates narrow corridors in subplots (a, b, c) in Figure \ref{fig:overview}.
Due to the conservatism of the OCR framework, a few adjustments are needed for good performance in highly cluttered settings.
First, we discard LiDAR readings outside of a front-facing cone spanning $\pi/2$ rad.
We believe this is necessary because of a shift in distribution of LiDAR readings from what is seen during training, which has sparse obstacles and no walls.
Second, we use the uncalibrated output of the OCR-VN.
We theorize that the conservatism of the OCR framework is an inherent result of using a worst-case analysis for disturbance, which makes the safety problem especially difficult in crowded settings.
We defer addressing these limitations to future work.

\subsubsection{Rough Terrain}

The robot avoids collisions on rough outdoor terrains in subplots (d, i, j) of Figure \ref{fig:overview}, further highlighting the ability of the proposed framework to adapt to dynamics uncertainties.

%% file: tables/hardware_results_table.tex
\begin{table*}[!t]
\caption{Hardware Results for OCR and ABS across Nominal Controllers (10 Trials)\label{tab:hardware_results}}
\centering
\begin{tabular*}{\linewidth}{@{\extracolsep{\fill}} ll@{\qquad\qquad}|ccc@{\qquad\qquad}|ccc }

\toprule
& & \multicolumn{3}{c|}{\textbf{Normal Condition}} & \multicolumn{3}{c}{\textbf{Slippery Condition}} \\
Controller & Filter & \multicolumn{1}{c}{\# Successes $\uparrow$} & \multicolumn{1}{c}{\# Collisions $\downarrow$} & \multicolumn{1}{c|}{\# Timeouts $\downarrow$} & \multicolumn{1}{c}{\# Successes $\uparrow$} & \multicolumn{1}{c}{\# Collisions $\downarrow$} & \multicolumn{1}{c}{\# Timeouts $\downarrow$} \\
\midrule
\multirow{3}{*}{ABS-Agile}       & No Filter           & 9  & 1  & 0 & 1 & 9  & 0 \\
                           & ABS                 & 9  & 1  & 0 & 2 & 8  & 0 \\
                           & \textbf{OCR (ours)} & \textbf{10} & \textbf{0}  & 0 & \textbf{8} & \textbf{1}  & 1 \\
\midrule
\multirow{2}{*}{PS + WTW}  & No Filter           & 5  & 5  & 0 & 1 & 9  & 0 \\
                           & \textbf{OCR (ours)} & \textbf{9}  & \textbf{1}  & 0 & \textbf{8} & \textbf{2}  & 0 \\
\midrule
\multirow{2}{*}{NVE + WTW} & No Filter           & 0  & 10 & 0 & 0 & 10 & 0 \\
                           & \textbf{OCR (ours)} & \textbf{9}  & \textbf{0}  & 1 & \textbf{8} & \textbf{2}  & 0 \\
\midrule
\multirow{2}{*}{PS + MPC}  & No Filter           & 5  & 5  & 0 & 0 & 10 & 0 \\
                           & \textbf{OCR (ours)} & \textbf{10} & \textbf{0}  & 0 & \textbf{9} & \textbf{1}  & 0 \\
\midrule
\multirow{2}{*}{NVE + MPC} & No Filter           & 0  & 10 & 0 & 0 & 10 & 0 \\
                           & \textbf{OCR (ours)} & \textbf{9}  & \textbf{0}  & 1 & \textbf{7} & \textbf{3}  & 0 \\
\bottomrule

\end{tabular*}
\end{table*}

%% file: figures/slippery.tex
\begin{figure*}[!t]
\centering
\includegraphics[width=1.0\textwidth]{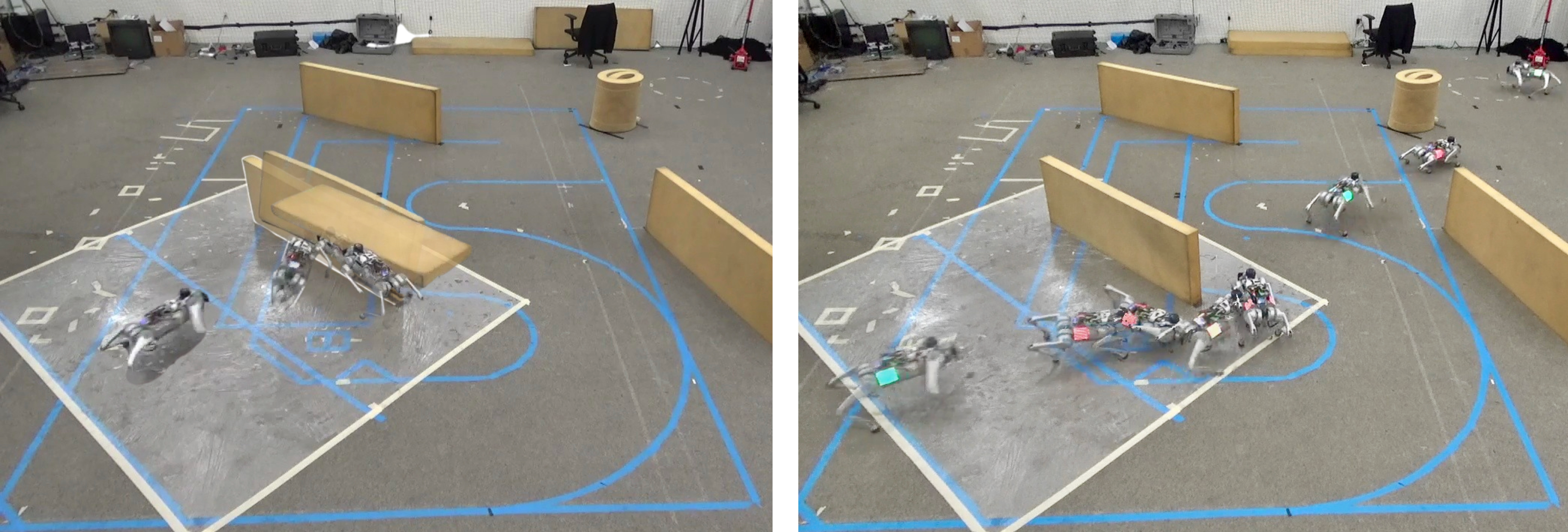}
\caption{Hardware experiments with a slippery region outlined in white. (Left) The ABS baseline collides due to drifting caused by the slippery floor. (Right) The OCR framework ({\color{ForestGreen}green}/{\color{BrickRed}red} : nominal/filtered) stops and turns in time to prevent a collision, demonstrating superior robustness to changes in the system dynamics.}
\label{fig:slippery}
\end{figure*}

%% file: sections/limitations.tex
\section{limitations}\label{sec:limitations}

Although our proposed framework achieves high safety rates in both simulation and hardware results in Tables \ref{tab:simulation_results} and \ref{tab:hardware_results}, the framework still exhibits a nonzero collision rate.
We attribute this to several failure modes of the framework that we identify and discuss next.

First, during deployment, the system may experience disturbances that exceed the estimated disturbance bounds which the framework is designed to safeguard against.
Furthermore, there is an unavoidable delay before environment changes will reflect in the estimated disturbance bounds.
During this delay period, the system can encounter failure before the framework has a chance to adapt.

Second, the OCR-VN can contain learning errors that are critical to safety.
While the calibration scheme presented in Section \ref{sec:calibration} can help us better understand the degree of these errors, we cannot directly extrapolate the theoretical guarantees from a validation dataset to the real world, due to potential distribution shifts.
It would be interesting to explore online calibration methods to overcome this challenge.

In addition to these failure modes which affect safety, there are several limitations of the framework that affect its performance.
For some low-level locomotion policies and environment settings, the error in the reduced-order dynamics model can be very large, leading to high dynamical uncertainty.
Large estimated disturbances can also appear as an artifact of latency and state estimation issues with real-world hardware.
The resulting disturbance bounds can produce overly conservative BRTs which severely inhibit progress by the system.
In highly cluttered environments, this can manifest as stalling behavior.

The conservatism of the framework is also partly a consequence of our choice to model the dynamics uncertainty as \textit{adversarial} in nature.
Future works can explore approaches to overcome these issues by using proactive methods of modeling and estimating dynamics and disturbances, for example by using observations of the environment or learning controller-specific dynamics to anticipate future system behavior and reduce uncertainty, as well as considering different characterizations of disturbances.

%% file: sections/conclusion.tex
\section{Conclusion}\label{sec:conclusion}

We propose the OCR framework, which uses an adaptive safety filter to robustly ensure the safety of a quadruped robot running an \textit{a priori unknown} locomotion policy in \textit{a priori unknown} environments using only LiDAR observations.
The offline training of the OCR-VN is done \textit{without a priori access to the controllers or a simulator} - hence, ``One Filter to Deploy Them All''.
In simulation and hardware experiments on a Unitree Go1 quadruped, we demonstrate the superior efficacy of the proposed approach to ensure, in zero-shot fashion, the safety of numerous controllers across diverse environment configurations and perturbed dynamics.
Videos demonstrating the efficacy of the proposed approach across various settings can be found on the project website listed above Figure \ref{fig:overview}.

%% file: appendices/proof.tex
\section{Proof of Theorem \ref{thm:calibration}}\label{apd:proof}

\begin{proof}

Theorem \ref{thm:calibration} is a straightforward application of the split conformal prediction method detailed in \cite{MAL-101}, where we set their ``input $x$'' as our network input $(x_r, \bar{d}^e_r, o^e)$, their ``output $y$'' as our true value $V\left( x_r; e \right)$, their ``score function $s(x,y)$'' as our prediction error $V_\psi (x_r, \bar{d}^e_r, o^e)-V\left( x_r; e \right)$, their ``calibration size $n$'' as our calibration size $N$, their ``error rate $\alpha$'' in terms of our number of outliers $k$ as $\frac{k+1}{N+1}$, and their ``quantile $\hat{q}$'' as our calibration level $\delta$.
Indeed, recall that we compute $\delta$ as the $\frac{N-k}{N}=\frac{N-(\alpha(N+1)-1)}{N}=\frac{(N+1)(1-\alpha)}{N}$ quantile of the calibration scores $\{s_i\}_{i=1}^N$, which is precisely how $\hat{q}$ is computed in \cite{MAL-101}.
Section 3.2 in \cite{MAL-101} yields:
\begin{equation}\label{eq:beta}
    \underset{ \left(x_r, \bar{d}^e_r, o^e\right) }{\mathbb{P}} \left( V_\psi (x_r, \bar{d}^e_r, o^e)-V\left( x_r; e \right) \le \delta  \right) \sim \mathcal{B}(N-k,k+1),
\end{equation}
where $\mathcal{B}$ is the Beta distribution.
Line \eqref{eq:beta} tells us that the probability that the OCR-VN overestimates the true value by at most $\delta$ follows a Beta distribution that depends on the number of calibration points $N$ and outliers $k$ used to determine $\delta$.
We are interested in lower bounding this probability by $1-\epsilon$ with a confidence of at least $1-\beta$.
Fortunately, since the relationship on Line \eqref{eq:beta} is known to us, we can compute the $k$ needed to satisfy a desired $\epsilon$ and $\beta$.
The cumulative distribution function of the Beta distribution on Line \eqref{eq:beta} evaluated at $x$ is given in terms of $k$ by the incomplete beta function ratio $I_x(N-k, k+1)=\sum^N_{j=N-k}\binom{N}{j}x^j(1-x)^{N-j}$ from \href{https://dlmf.nist.gov/8.17.E5}{DLMF, (8.17.5)} \cite{NIST:DLMF}.
Changing the index $i=N-j$ yields $I_x(N-k, k+1)=\sum^k_{i=0}\binom{N}{N-i}x^{N-i}(1-x)^{N-(N-i)}=\sum^k_{i=0}\binom{N}{i}x^{N-i}(1-x)^{i}$.
Thus, Line \eqref{eq:beta} is equivalent to the claim that for any violation parameter $\epsilon \in (0, 1)$ and confidence parameter $\beta \in (0, 1)$, $\underset{ \left(x_r, \bar{d}^e_r, o^e\right) }{\mathbb{P}} \left( V_\psi (x_r, \bar{d}^e_r, o^e)-V\left( x_r; e \right) \le \delta  \right) \ge 1-\epsilon$ holds with probability at least $1-\beta$ over the draws of the samples as long as $\beta \ge I_{1-\epsilon}(N-k, k+1)=\sum^k_{i=0}\binom{N}{i}\epsilon^{i}(1-\epsilon)^{N-i}$.
This is the same requirement on $k$ as presented in Theorem \ref{thm:calibration}.

\end{proof}

%% file: appendices/results.tex
\section{All Experiment Results}\label{apd:results}
Simulation experiments are conducted in easy, medium, and hard conditions to determine the effect of increasing dynamical variation on framework efficacy.
The experiment difficulty determines the range of possible payload and friction parameters, as listed in Table \ref{tab:simulation_parameters}.
As the difficulty increases, the parameter ranges stray further from normal conditions.
We report results for the hard condition in Section \ref{sec:simulation_experiments} in the main text.
All simulation results can be found in Table \ref{tab:simulation_all_results}.

All hardware results can be found in Table \ref{tab:hardware_all_results}.

\FloatBarrier
\input{tables/simulation_parameters_table}
\input{tables/simulation_all_results_table}
\input{tables/hardware_all_results_table}
\FloatBarrier

%% file: tables/simulation_parameters_table.tex
\begin{table*}[!t]
\caption{Simulation Environment Parameters\label{tab:simulation_parameters}}
\centering
\begin{tabular*}{\linewidth}{@{\extracolsep{\fill}} ccc|ccc }

\toprule
\multicolumn{3}{c|}{\textbf{Payload Range (kg)}} & \multicolumn{3}{c}{\textbf{Friction Range}} \\
\midrule
Easy & Medium & Hard & Easy & Medium & Hard \\
$[0, 0]$ & $[-1, 1]$ & $[-1, -0.5] \cup [0.5, 1]$ & $[1, 1]$ & $[0.5, 1.5]$ & $[0.5, 0.75] \cup [1.25, 1.5]$ \\
\bottomrule

\end{tabular*}
\end{table*}

%% file: tables/simulation_all_results_table.tex
\begin{table*}[!t]
\caption{All Simulation Results (100 Trials)\label{tab:simulation_all_results}}
\centering
\begin{tabular*}{\linewidth}{@{\extracolsep{\fill}} l|ll@{\qquad\qquad}|cccccc }

\toprule
\textbf{Difficulty} & \textbf{Controller} & \textbf{Filter} & \textbf{Success Rate $\uparrow$} & \textbf{Collision Rate $\downarrow$} & \textbf{Timeout Rate $\downarrow$} & $\mathbf{\bar{v}}$ \textbf{(m/s) $\uparrow$} & $\mathbf{\bar{r} \downarrow}$ & $\mathbf{\bar{q}}$ \textbf{(m) $\uparrow$} \\
\midrule
\multirow{12}{*}{Easy}
    & \multirow{5}{*}{ABS-Agile}
        & No Filter                               & 0.89 & 0.11 & 0.00 & \textbf{2.12} & \textbf{0.00} & 0.44 \\
    &   & ABS Framework & 0.88 & 0.12 & 0.00 & 2.08 & 0.02 & 0.44 \\
    &   & OCR $\backslash$ DE & 0.51 & 0.21 & 0.28 & 0.86 & 0.68 & 0.44 \\
    &   & OCR $\backslash$ C            & 0.88 & 0.12 & 0.00 & 1.77 & 0.26 & 0.46 \\
    &   & \textbf{OCR (ours)}                     & \textbf{0.95} & \textbf{0.05} & 0.00 & 1.24 & 0.60 & \textbf{0.56} \\ \cmidrule{2-9}
    & \multirow{4}{*}{PS + WTW}
        & No Filter                               & 0.80 & 0.20 & 0.00 & \textbf{1.46} & \textbf{0.00} & 0.44 \\
    &   & OCR $\backslash$ DE & 0.85 & 0.15 & 0.00 & 0.84 & 0.62 & 0.43 \\
    &   & OCR $\backslash$ C            & 0.91 & 0.09 & 0.00 & 1.31 & 0.14 & 0.42 \\
    &   & \textbf{OCR (ours)}                     & \textbf{1.00} & \textbf{0.00} & 0.00 & 0.99 & 0.46 & \textbf{0.62} \\ \cmidrule{2-9}
    & \multirow{4}{*}{NVE + WTW}
        & No Filter                               & 0.25 & 0.75 & 0.00 & \textbf{1.82} & \textbf{0.00} & 0.36 \\
    &   & OCR $\backslash$ DE & 0.22 & 0.78 & 0.00 & 1.23 & 0.50 & 0.40 \\
    &   & OCR $\backslash$ C            & 0.46 & 0.54 & 0.00 & 1.53 & 0.22 & 0.37 \\
    &   & \textbf{OCR (ours)}                     & \textbf{0.98} & \textbf{0.02} & 0.00 & 1.06 & 0.48 & \textbf{0.58} \\
\midrule
\multirow{12}{*}{Medium}
    & \multirow{5}{*}{ABS-Agile}
        & No Filter                               & 0.83 & 0.17 & 0.00 & \textbf{2.12} & \textbf{0.00} & 0.43 \\
    &   & ABS Framework & 0.87 & 0.13 & 0.00 & 2.05 & 0.02 & 0.43 \\
    &   & OCR $\backslash$ DE & 0.53 & 0.27 & 0.20 & 1.02 & 0.61 & 0.45 \\
    &   & OCR $\backslash$ C            & 0.84 & 0.16 & 0.00 & 1.74 & 0.27 & 0.44 \\
    &   & \textbf{OCR (ours)}                     & \textbf{0.93} & \textbf{0.07} & 0.00 & 1.27 & 0.57 & \textbf{0.55} \\ \cmidrule{2-9}
    & \multirow{4}{*}{PS + WTW}
        & No Filter                               & 0.81 & 0.19 & 0.00 & \textbf{1.42} & \textbf{0.00} & 0.46 \\
    &   & OCR $\backslash$ DE & 0.81 & 0.19 & 0.00 & 0.86 & 0.61 & 0.44 \\
    &   & OCR $\backslash$ C            & 0.83 & 0.17 & 0.00 & 1.33 & 0.12 & 0.46 \\
    &   & \textbf{OCR (ours)}                     & \textbf{0.99} & \textbf{0.01} & 0.00 & 1.00 & 0.42 & \textbf{0.65} \\ \cmidrule{2-9}
    & \multirow{4}{*}{NVE + WTW}
        & No Filter                               & 0.25 & 0.75 & 0.00 & \textbf{1.79} & \textbf{0.00} & 0.39 \\
    &   & OCR $\backslash$ DE & 0.22 & 0.78 & 0.00 & 1.11 & 0.56 & 0.40 \\
    &   & OCR $\backslash$ C            & 0.42 & 0.58 & 0.00 & 1.47 & 0.22 & 0.39 \\
    &   & \textbf{OCR (ours)}                     & \textbf{0.97} & \textbf{0.03} & 0.00 & 1.05 & 0.50 & \textbf{0.57} \\
\midrule
\multirow{12}{*}{Hard}
    & \multirow{5}{*}{ABS-Agile}
        & No Filter                               & 0.75 & 0.25 & 0.00 & \textbf{2.10} & \textbf{0.00} & 0.41 \\
    &   & ABS Framework & 0.80 & 0.20 & 0.00 & 2.03 & 0.03 & 0.41 \\
    &   & OCR $\backslash$ DE & 0.40 & 0.35 & 0.25 & 0.98 & 0.62 & 0.41 \\
    &   & OCR $\backslash$ C            & 0.81 & 0.19 & 0.00 & 1.70 & 0.28 & 0.41 \\
    &   & \textbf{OCR (ours)}                     & \textbf{0.91} & \textbf{0.09} & 0.00 & 1.22 & 0.59 & \textbf{0.58} \\ \cmidrule{2-9}
    & \multirow{4}{*}{PS + WTW}
        & No Filter                               & 0.72 & 0.28 & 0.00 & \textbf{1.36} & \textbf{0.00} & 0.43 \\
    &   & OCR $\backslash$ DE & 0.78 & 0.22 & 0.00 & 0.84 & 0.61 & 0.42 \\
    &   & OCR $\backslash$ C            & 0.90 & 0.09 & 0.01 & 1.21 & 0.18 & 0.44 \\
    &   & \textbf{OCR (ours)}                     & \textbf{1.00} & \textbf{0.00} & 0.00 & 0.97 & 0.42 & \textbf{0.66} \\ \cmidrule{2-9}
    & \multirow{4}{*}{NVE + WTW}
        & No Filter                               & 0.20 & 0.80 & 0.00 & \textbf{1.70} & \textbf{0.00} & 0.31 \\
    &   & OCR $\backslash$ DE & 0.21 & 0.79 & 0.00 & 0.93 & 0.61 & 0.37 \\
    &   & OCR $\backslash$ C            & 0.45 & 0.55 & 0.00 & 1.32 & 0.28 & 0.39 \\
    &   & \textbf{OCR (ours)}                     & \textbf{0.91} & \textbf{0.08} & 0.01 & 1.04 & 0.47 & \textbf{0.62} \\
\bottomrule

\end{tabular*}
\end{table*}

%% file: tables/hardware_all_results_table.tex
\begin{table*}[!t]
\caption{All Hardware Results (10 Trials)\label{tab:hardware_all_results}}
\centering
\begin{tabular*}{\linewidth}{@{\extracolsep{\fill}} l|ll@{\qquad\qquad}|cccccc }

\toprule
\textbf{Condition} & \textbf{Controller} & \textbf{Filter} & \textbf{\# Successes $\uparrow$} & \textbf{\# Collisions $\downarrow$} & \textbf{\# Timeouts $\downarrow$} & $\mathbf{\bar{v}}$ \textbf{(m/s) $\uparrow$} & $\mathbf{\bar{r} \downarrow}$ & $\mathbf{\bar{q}}$ \textbf{(m) $\uparrow$} \\
\midrule
\multirow{11}{*}{Normal}
    & \multirow{3}{*}{ABS-Agile}
        & No Filter                               & 9 & 1 & 0 & \textbf{1.67} & \textbf{0.00} & 0.29 \\
    &   & ABS                                     & 9 & 1 & 0 & 1.59 & 0.02 & 0.29 \\
    &   & \textbf{OCR (ours)}                     & \textbf{10} & \textbf{0} & 0 & 0.94 & 0.64 & \textbf{0.43} \\ \cmidrule{2-9}
    & \multirow{2}{*}{PS + WTW}
        & No Filter                               & 5 & 5 & 0 & \textbf{1.53} & \textbf{0.00} & 0.32 \\
    &   & \textbf{OCR (ours)}                     & \textbf{9} & \textbf{1} & 0 & 0.80 & 0.58 & \textbf{0.61} \\ \cmidrule{2-9}
    & \multirow{2}{*}{NVE + WTW}
        & No Filter                               & 0 & 10 & 0 & N/A & N/A & N/A \\
    &   & \textbf{OCR (ours)}                     & \textbf{9} & \textbf{0} & 1 & \textbf{0.77} & \textbf{0.61} & \textbf{0.55} \\ \cmidrule{2-9}
    & \multirow{2}{*}{PS + MPC}
        & No Filter                               & 5 & 5 & 0 & \textbf{1.55} & \textbf{0.00} & 0.33 \\
    &   & \textbf{OCR (ours)}                     & \textbf{10} & \textbf{0} & 0 & 0.81 & 0.54 & \textbf{0.58} \\ \cmidrule{2-9}
    & \multirow{2}{*}{NVE + MPC}
        & No Filter                               & 0 & 10 & 0 & N/A & N/A & N/A \\
    &   & \textbf{OCR (ours)}                     & \textbf{9} & \textbf{0} & 1 & \textbf{0.93} & \textbf{0.56} & \textbf{0.54} \\
\midrule
\multirow{11}{*}{Slippery}
    & \multirow{3}{*}{ABS-Agile}
        & No Filter                               & 1 & 9 & 0 & 1.38 & \textbf{0.00} & 0.33 \\
    &   & ABS                                     & 2 & 8 & 0 & \textbf{1.49} & 0.03 & 0.22 \\
    &   & \textbf{OCR (ours)}                     & \textbf{8} & \textbf{1} & 1 & 0.90 & 0.60 & \textbf{0.38} \\ \cmidrule{2-9}
    & \multirow{2}{*}{PS + WTW}
        & No Filter                               & 1 & 9 & 0 & \textbf{1.04} & \textbf{0.00} & \textbf{0.52} \\
    &   & \textbf{OCR (ours)}                     & \textbf{8} & \textbf{2} & 0 & 0.74 & 0.65 & 0.45 \\ \cmidrule{2-9}
    & \multirow{2}{*}{NVE + WTW}
        & No Filter                               & 0 & 10 & 0 & N/A & N/A & N/A \\
    &   & \textbf{OCR (ours)}                     & \textbf{8} & \textbf{2} & 0 & \textbf{0.61} & \textbf{0.68} & \textbf{0.39} \\ \cmidrule{2-9}
    & \multirow{2}{*}{PS + MPC}
        & No Filter                               & 0 & 10 & 0 & N/A & N/A & N/A \\
    &   & \textbf{OCR (ours)}                     & \textbf{9} & \textbf{1} & 0 & \textbf{0.80} & \textbf{0.54} & \textbf{0.56} \\ \cmidrule{2-9}
    & \multirow{2}{*}{NVE + MPC}
        & No Filter                               & 0 & 10 & 0 & N/A & N/A & N/A \\
    &   & \textbf{OCR (ours)}                     & \textbf{7} & \textbf{3} & 0 & \textbf{0.82} & \textbf{0.57} & \textbf{0.32} \\
\bottomrule

\end{tabular*}
\end{table*}